\documentclass[10pt,twocolumn,letterpaper]{article}

\makeatletter
\@namedef{ver@everyshi.sty}{}
\makeatother

\usepackage{cvpr}
\usepackage{times}
\usepackage{epsfig}
\usepackage{graphicx}
\usepackage{amsmath}
\usepackage{amssymb}

\usepackage[pagebackref=true,breaklinks=true,letterpaper=true,colorlinks,bookmarks=false]{hyperref}

\usepackage{pgfplots}
\usepackage{tikz}
\usepackage[utf8]{inputenc} % allow utf-8 input
\usepackage[T1]{fontenc}    % use 8-bit T1 fonts
\usepackage{url}            % simple URL typesetting
\usepackage{booktabs}       % professional-quality tables
\usepackage{amsfonts}       % blackboard math symbols
\usepackage{nicefrac}       % compact symbols for 1/2, etc.
\usepackage{microtype}      % microtypography
\usepackage{multirow}
\usepackage{graphicx}
\usepackage{caption}
\usepackage{subcaption}
\usepackage{color,soul}
\usepackage{array}
\usepackage{enumitem}
\usepackage{adjustbox}
\usepackage{wrapfig}
\newcolumntype{$}{>{\global\let\currentrowstyle\relax}}
\newcolumntype{^}{>{\currentrowstyle}}

\usepackage{bm}

\usepackage{xcolor}
\usepackage{bbding}
\usepackage{pifont}
\newcommand{\cmark}{\color{green} \ding{51}}%
\newcommand{\xmark}{\color{red} \ding{55}}%
\newcolumntype{R}[2]{%
    >{\adjustbox{angle=#1,lap=\width-(#2)}\bgroup}%
    l%
    <{\egroup}%
}
\newcommand*\rot{\multicolumn{1}{R{45}{1em}}}% no optional argument here, please!

\newcommand{\var}{\text{Var}}
\newcommand{\cov}{\text{Cov}}

\newcommand{\z}{{\pmb{z}}}

\newcommand{\train}{\text{train}}
\newcommand{\val}{\text{val}}
\newcommand{\gen}{\text{gen}}

\renewcommand{\i}{{\pmb{i}}}
\renewcommand{\o}{{\pmb{o}}}

\newcommand{\w}{{\pmb{w}}}

\newcommand{\bzeta}{{\pmb{\zeta}}}

\newcommand{\bphi}{{\pmb{\phi}}}
\renewcommand{\L}{{\mathcal{L}}}
\newcommand{\Ld}{{\mathcal{L}_d}}
\newcommand{\Ln}{{\mathcal{L}_n}}

\newcommand{\E}{{\mathbb{E}}}

\cvprfinalcopy % *** Uncomment this line for the final submission

\def\cvprPaperID{7848} % *** Enter the CVPR Paper ID here

\ifcvprfinal\fi
\begin{document}

%%%%%%%%% TITLE
\title{UNAS: Differentiable Architecture Search Meets Reinforcement Learning}

\author{Arash Vahdat, Arun Mallya, Ming-Yu Liu, Jan Kautz \\
NVIDIA \\
{\tt\small \{avahdat, amallya, mingyul, jkautz\}@nvidia.com}
% \and
}

\maketitle
%\thispagestyle{empty}

%!TEX root = main.tex
\begin{abstract}
Neural architecture search (NAS) aims to discover network architectures with desired properties such as high accuracy or low latency. Recently, differentiable NAS (DNAS) has demonstrated promising results while maintaining a search cost orders of magnitude lower than reinforcement learning (RL) based NAS. However, DNAS models can only optimize differentiable loss functions in search, and they require an accurate differentiable approximation of non-differentiable criteria. In this work, we present UNAS, a unified framework for NAS, that encapsulates recent DNAS and RL-based approaches under one framework. Our framework brings the best of both worlds, and it enables us to search for architectures with both differentiable and non-differentiable criteria in one unified framework while maintaining a low search cost. Further, we introduce a new objective function for search based on the generalization gap that prevents the selection of architectures prone to overfitting. We present extensive experiments on the CIFAR-10, CIFAR-100 and ImageNet datasets and we perform search in two fundamentally different search spaces. We show that UNAS obtains the state-of-the-art average accuracy on all three datasets when compared to the architectures searched in the DARTS~\cite{liu2018darts} space. Moreover, we show that UNAS can find an efficient and accurate architecture in the ProxylessNAS~\cite{sandler2018mobilenetv2} search space, that outperforms existing MobileNetV2~\cite{sandler2018mobilenetv2} based architectures. The source code is available at \url{https://github.com/NVlabs/unas}.
\end{abstract}

%!TEX root = main.tex
\vspace{-0.2cm}
\section{Introduction}

Since the success of deep learning, designing neural network architectures with desirable performance criteria (\eg high accuracy, low latency, \etc) for a given task has been a challenging problem. Some call it alchemy and some refer to it as intuition, but the task of discovering a novel architecture 
often involves a tedious and costly process of trial-and-error for searching in an exponentially large space of hyper-parameters.
The goal of neural architecture search (NAS)~\cite{elsken2018neural} is to find novel networks for new problem domains and criteria automatically and efficiently.

Early work on NAS used reinforcement learning~\cite{baker2016designing,cai2018efficient,pham2018efficient,zoph2016neural,zoph2018learning}, or evolutionary algorithms~\cite{liu2017hierarchical,real2017large,real2018regularized,xie2017genetic} to obtain state-of-the-art performance on a variety of tasks. Although, these methods are generic and can search for architecture with a broad range of criteria, they are often computationally demanding. For example, the RL-based approach~\cite{zoph2018learning},
and evolutionary method~\cite{real2018regularized} each requires over 2000 GPU days. 

Recently, several differentiable neural architecture search (DNAS) frameworks~\cite{liu2018darts, xie2018snas, wu2018fbnet, cai2018proxylessnas} have shown promising results while reducing the search cost to a few GPU days. However, these approaches assume that the objective function is differentiable with respect to the architecture parameters and cannot directly optimize non-differentiable criteria like network latency, power consumption, memory usage, \etc. To tackle this problem, DNAS methods~\cite{wu2018fbnet, cai2018proxylessnas, yang2018netadapt} approximate network latency using differentiable functions. However, these approximations may fail when the underlying criteria cannot be accurately modeled.
For example, if compiler optimizations are used, methods such as layer fusion, mixed-precision inference, and kernel auto-tuning can dramatically change latency, making it challenging to approximate it accurately. In addition to the loss approximation, DNAS relies on the continuous approximation of discrete variables in search, introducing additional mismatch in network performance between discovered architecture and the corresponding continuous relaxations.

In this paper, we introduce UNAS, a unified framework for NAS that bridges the gap between DNAS and RL-based architecture search. {\bf (i)} UNAS offers the best of both worlds and enables us to search for architectures using both differentiable objective functions (\eg, cross-entropy loss) and non-differentiable functions (\eg, network latency). UNAS keeps the search time low similar to other DNAS models, but it also eliminates the need for accurate approximation of non-differentiable criteria. {\bf (ii)} UNAS training does not introduce any additional biases due to the continuous relaxation of architecture parameters. We show that the gradient estimation in UNAS is equal to the estimations obtained by RL-based frameworks that operate on discrete variables.
%{\bf (ii)} trading of gradient noise for search efficiency (in terms of GPU memory and hours) without introducing any additional bias, 
Finally, {\bf (iii)} UNAS proposes a new objective function based on the generalization gap which is empirically shown to find architectures less prone to overfitting. 

We perform extensive experiments in both DARTS~\cite{liu2018darts} and ProxylessNAS~\cite{cai2018proxylessnas} search spaces. We show that UNAS achieves the state-of-the-art average performance on all three datasets in comparison to the recent gradient-based NAS models in the DARTS space. Moreover, UNAS can find architectures that are faster and more accurate than architectures, searched in the ProxylessNAS space.

%!TEX root = main.tex

\subsection{Related Work}\label{sec:related}
%{\bf Related Work:} 
Zoph and Le~\cite{zoph2016neural} introduced the paradigm of NAS, where a controller recurrent neural network (RNN) was trained to output the specification of a network (filter sizes, number of channels, \etc). The controller was trained using REINFORCE~\cite{williams1992simple} to maximize the expected accuracy of the output network on the target validation set, after training on the target task. 
Requiring the method to specify every layer of the network 
made it challenging to deepen or transfer an obtained network to other tasks. 
Based on the observation that popular manually-designed convolutional neural networks (CNNs) such as ResNet~\cite{he2016deep} or Inception~\cite{szegedy2015going} contained repeated generic blocks with the same structure,
Zoph \etal~\cite{zoph2018learning}
trained the RNN to output stackable `\emph{cells}'. The task of NAS was thus reduced to learning two types of cells, the {\it Normal Cell} - convolutional cells that preserve the spatial dimensions, and the {\it Reduce Cell} - convolutional cells that reduce spatial dimensions while increasing feature maps.
% \arun{two inputs per node}

Recently, DARTS~\cite{liu2018darts} relaxed the architecture search space to be continuous by using a weighted mixture-of-operations and optimized the candidate architecture through gradient descent. 
Using weight-sharing~\cite{bender2018understanding,pham2018efficient}, they brought search down to a few GPU days.
As the final architecture is required to be discrete, 
DARTS only retained the top two operations based on the weight assigned to each operation.
Building upon DARTS, SNAS~\cite{xie2018snas} used weights sampled from a trainable Gumbel-Softmax distribution instead of continuous weights.
Both DARTS and SNAS assume that the objective function for search is differentiable. We extend these frameworks by introducing unbiased gradient estimators that can work for both differentiable and non-differentiable objective functions.

Recent works~\cite{cai2018proxylessnas, wu2018fbnet, yang2018netadapt, Howard2019Mobilenetv3, tan2018mnasnet} consider latency in architectures search. 
ProxylessNAS~\cite{cai2018proxylessnas}, FBNet~\cite{wu2018fbnet} and NetAdapt~\cite{yang2018netadapt} convert the non-differentiable latency objective to a differentiable function by learning an accurate latency approximation. However, these approximations may fail when latency cannot be predicted by a trainable function.
MnasNet~\cite{tan2018mnasnet} does not require a differentiable approximation of the latency as it relies on an RL-objective, however, it requires $\sim$300 TPU-days for each architecture search. Our framework bridges the gap between differentiable and RL-based NAS; it can search with differentiable and non-differentiable functions and it does not require an accurate approximation of non-differentiable terms in the objective. Our work is compared against previous works in Table.~\ref{table:comparison}.

Recently, P-DARTS~\cite{chen2019pdarts} proposes a progressive version of DARTS and shows that by gradually increasing the depth of the network during the search, deeper cells can be discovered. UNAS explores an orthogonal direction to P-DARTS and it proposes generic gradient estimators that work with both differentiable and non-differentiable losses and new generalization-based search objective functions. 

\begin{table}[]
\setlength{\tabcolsep}{3pt}
    % \centering
    \begin{tabular}{lc|c|c|c|c|c|c}
        & \rot{\small  DARTS~\cite{liu2018darts}} 
        & \rot{\small  SNAS~\cite{xie2018snas}} 
        & \rot{\small  P-DARTS~\cite{liu2018progressive}} 
        & \rot{\small  ProxylessNAS~\cite{cai2018proxylessnas}} 
        & \rot{\small FBNet~\cite{wu2018fbnet}} 
        & \rot{\small MnasNet~\cite{tan2018mnasnet}} 
        & \rot{\small UNAS (ours)}\\
        %  \multicolumn{1}{l|}{Differentiable Cell Search} & \ding{51} & \ding{51} & \ding{51}  \\
         \multicolumn{1}{l|}{Differentiable loss} & \cmark & \cmark & \cmark & \cmark & \cmark & \cmark & \cmark \\
         \multicolumn{1}{l|}{Non-differentiable loss} & \xmark & \xmark & \xmark &  \xmark &  \xmark & \cmark & \cmark \\
         \multicolumn{1}{l|}{Latency optimization} & \xmark & \xmark & \xmark & \cmark & \cmark & \cmark & \cmark \\
         \multicolumn{1}{l|}{Low search cost} & \cmark & \cmark & \cmark & \cmark & \cmark & \xmark & \cmark \\
         %\multicolumn{1}{l|}{Low variance} & \xmark & \xmark & \xmark & \xmark & \cmark
    \end{tabular}
    \caption{Comparison with differentiable NAS methods.}
    \label{table:comparison}
    \vspace{-0.5cm}
\end{table}

%!TEX root = main.tex
\vspace*{-0.1cm}
\section{Background} 
\label{sec:background}
In differentiable architecture search (DARTS)~\cite{liu2018darts}, a network is represented by a directed acyclic graph, where each node in the graph denotes a hidden representation (\eg, feature maps in CNNs) and each directed edge represents an operation transforming the state of the input node. The n$^\mathit{th}$ node $x_n$ is connected to its predecessors (\ie, $P_n$) and its content is computed by applying a set of operations to the predeceasing nodes, represented by $ x_n = \sum_{x_m \in P_n} O_{m, n}(x_m)$,
%\begin{equation*}
%    x_n = \sum_{m<n} O_{m, n}(x_m),
%\end{equation*}
where $O_{m, n}$ is the operation applied to $x_m$. The goal of architecture search is then to find the operation $O_{m, n}$ for each edge $(m, n)$. Representing the set of all possible operations that can be applied to the edge $e:=(m, n)$ using $\{O_{e}^{(1)}, O_{e}^{(2)}, \dots, O_{e}^{(K)}\}$ where $K$ is the number of operations, this discrete assignment problem can be formulated as a \textit{mixed operation} denoted by $O_{e}(x_m) = \sum_{k=1}^{K} z_{e}^{(k)} O_{e}^{(k)}(x_m)$,
%\begin{equation*}
%    O_{e}(x_m) = \sum_{k=1}^{K} z_{e}^{(k)} O_{e}^{(k)}(x_m),
%\end{equation*}
where $\z_{e} = [z_{e}^{(1)}, z_{e}^{(2)}, \dots, z_{e}^{(K)}]$ is a one-hot binary vector (\ie, $z_{e}^{(k)} \in \{0, 1\}$) with a single one indicating the selected operation. Typically, it is assumed that the set of operations also includes a \textit{zero} operation that enables omitting edges in the network, and thus, learning the connectivity as well. 

We can construct a network architecture given the set of all operation assignments for all edges denoted by $\z = \{\z_e\}$. Therefore, the objective of the architecture search is to find a distribution over architecture parameters, $\z$ such that it minimizes the expected loss $\E_{p_{\bphi}(\z)}[\L(\z)]$ where $p_{\bphi}$ is a $\bphi$-parameterized distribution over $\z$ and $\L(\z)$ is a loss function measuring the performance of the architecture specified by $\z$ using a performance measure such as classification loss.

We assume that the architecture distribution is a factorial distribution with the form $p_{\bphi}(\z)=\prod_e p_{\bphi_e}(\z_e)$ where $p_{\bphi_e}(\z_e)$ is a  $\bphi_e$-parameterized categorical distribution defined over the one-hot vector $\z_e$.
Recently, SNAS~\cite{xie2018snas} proposed using the Gumbel-Softmax relaxation~\cite{maddison2016concrete, jang2016gumbelsoftmax} for optimizing the expected loss. In this case, the categorical distribution $p_{\bphi}(\z)$ is replaced with a Gumbel-Softmax distribution $p_{\bphi}(\bzeta)$ where $\bzeta$ denotes the continuous relaxation of the architecture parameter $\z$. SNAS assumes that the loss $\L(\z)$ is differentiable with respect to $\z$ and it uses the reparameterization trick to minimize the expectation of the relaxed loss $\E_{p_{\bphi}(\bzeta)}[\L(\bzeta)]$ instead of $\E_{p_{\bphi}(\z)}[\L(\z)]$.

\vspace{-0.2cm}
\section{Method}
As discussed above, the problem of NAS can be formulated as optimizing the expected loss $\E_{p_{\bphi}(\z)}[\L(\z)]$. In this section, we present our framework in two parts. In Sec.~\ref{sec:grad}, we start by presenting a general framework for computing $\frac{\partial}{\partial \bphi} \E_{p_{\bphi}(\z)}[\L(\z)]$ which is required for optimizing the expected loss. Then, we present our formulation of the loss function $\L(\z)$ in Sec.~\ref{sec:objective}. 

\pgfplotsset{ every non boxed x axis/.append style={x axis line style=-},
     every non boxed y axis/.append style={y axis line style=-}}

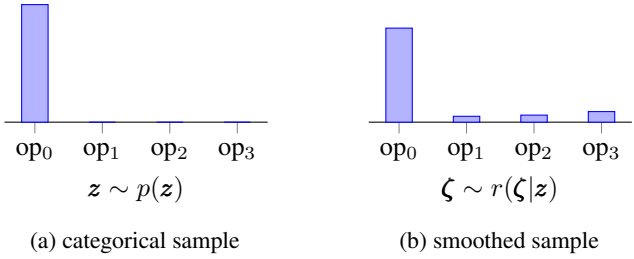
\begin{figure}
\vspace{-0.5cm}
\begin{subfigure}[b]{.2\textwidth}
\begin{tikzpicture}
\begin{axis}[
    name=mygraph,
    width=2.0in, height=1.3in, 
    axis x line=bottom, ymajorticks=false,
    ybar, enlarge x limits=0.15,
    y axis line style={draw opacity=0},
    symbolic x coords={op$_0$,op$_1$,op$_2$,op$_3$},
    xtick=data, %nodes near coords,
    ymin=0, ymax=1.1,
    ]
\addplot coordinates {(op$_0$,1) (op$_1$,0) (op$_2$,0) (op$_3$,0)};
\end{axis}
\node[anchor=north, yshift=-0.6cm] at (mygraph.south) {$\z \sim p(\z)$};
\end{tikzpicture}
\caption{categorical sample}
\end{subfigure} \hfill
\begin{subfigure}[b]{.2\textwidth}
\begin{tikzpicture}
\begin{axis}[
    name=mygraph,
    width=2.0in, height=1.3in, 
    axis x line=bottom, ymajorticks=false,
    ybar, enlarge x limits=0.15,
    y axis line style={draw opacity=0},
    symbolic x coords={op$_0$,op$_1$,op$_2$,op$_3$},
    xtick=data, %nodes near coords,
    ymin=0, ymax=1.1,
    ]
\addplot coordinates {(op$_0$,0.8) (op$_1$,0.05) (op$_2$,0.06) (op$_3$,0.09)};
\end{axis}
\node[anchor=north, yshift=-0.6cm] at (mygraph.south) {$\bzeta \sim r(\bzeta|\z)$};
\end{tikzpicture}
\caption{smoothed sample}
\end{subfigure}
\caption{(a) Operation selection corresponds to sampling from a categorical distribution that selects an operation. (b) Sampling from the conditional Gumbel-Softmax distribution $r(\bzeta|\z)$ acts as a smoothing distribution that yields continuous samples ($\bzeta$), correlated with the discrete samples ($\z$).} \label{fig:cond_sample}
\end{figure}

\subsection{Gradient Estimation}\label{sec:grad}
The most generic approach for optimizing the expected loss is the REINFORCE gradient estimator
\begin{equation} \label{eq:reinforce}
    \frac{\partial}{\partial \bphi} \E_{p_{\bphi}(\z)}[\L(\z)] = \E_{p_{\bphi}(\z)}\left[ \L(\z) \partial_\bphi \log p_{\bphi}(\z) \right],
\end{equation}
where $\partial \log p_{\bphi}(\z)$ is known as the score function and $\L(\z)$ is a loss function. As we can see, the gradient estimator in Eq.~\ref{eq:reinforce} only requires computing the loss function $\L(\z)$ (not the gradient $\partial_{\z} \L(\z)$), so it can be applied to any differentiable and non-differentiable loss function. However, this estimator is known to suffer from high variance and therefore a large number of trained architecture samples are required to reduce its variance, making it extremely compute intensive. 
% This typically translates to training many architecture samples which can be extremely compute intensive. 
The REINFORCE estimator in Eq.~\ref{eq:reinforce} can be also rewritten as
\begin{align} \label{eq:reinforce_base}
    \frac{\partial}{\partial \bphi} \E_{p_{\bphi}(\z)}[\L(\z)] = \ & \E_{p_{\bphi}(\z)}[\left(\L(\z) - c(\z)\right) \partial_\bphi \log  p_{\bphi}(\z)] \nonumber + \\ 
    & \ \partial_\bphi \E_{p_{\bphi}(\z)}[c(\z)],
\end{align}
where $c(\z)$ is a control variate~\cite{mcbook}. The gradient estimator in Eq.~\ref{eq:reinforce_base} has lower variance than Eq.~\ref{eq:reinforce}, if $c(\z)$ is correlated with $\L(\z)$,
and $\partial_\bphi \E_{p_{\bphi}(\z)}[c(\z)]$ has a low-variance gradient estimator~\cite{mnih2014neural,paisley2012variational, ranganath2014black}.\footnote{The low variance of Eq.~\ref{eq:reinforce_base} comes from fact that $\var(X - Y) = \var(X) + \var(Y) - 2\cov(X, Y)$ for any random variable $X$ and $Y$. If $X$ and $Y$ are highly correlated the negative contribution from $- 2\cov(X, Y)$ reduces the overall variance of $X - Y$.} 
% In the following, 
Without loss of generality, we assume that the loss function is decomposed into $\L(\z)\!=\!\Ld(\z)\!+\!\Ln(\z)$ where $\Ld(\z)$ contains the terms that are differentiable with respect to $\z$ and $\Ln(\z)$ includes the non-differentiable terms. We present a baseline function $c(\z)\!=\!c_d(\z)\!+\!c_n(\z)$, where $c_d(\z)$ and $c_n(\z)$ are for $\Ld(\z)$ and $\Ln(\z)$ respectively. Intuitively, the baseline is designed such that the term $\partial_\bphi \E_{p_{\bphi}(\z)}[c(\z)]$ in Eq.~\ref{eq:reinforce_base} is approximated using the low-variance reparameterization trick.

%For example, $\Ld(\z)$ can be the cross-entropy loss at the end of training network parameters or the loss at the current network parameters for the architecture specified by $\z$, and $\Ln(\z)$ can be the latency of the network.
{\bf Gradient Estimation for Differentiable Loss $\Ld$:} Following REBAR~\cite{tucker2017rebar}, in order to construct $c_d(\z)$, a control variate for $\Ld$, we use stochastic continuous relaxation $r_{\bphi}(\bzeta|\z)$ that samples from a conditional Gumbel-Softmax distribution given the architecture sample $\z$. Here, $\bzeta$\ can be considered as a smooth architecture defined based on $\z$ as shown in Fig.~\ref{fig:cond_sample}. Hence, it is highly correlated with $\z$ (see REBAR~\cite{tucker2017rebar} for details). With the definition $c_d(\z) := \E_{r_{\bphi}(\bzeta|\z)}[\Ld(\bzeta)]$, the gradient in Eq.~\ref{eq:reinforce_base} can be written as

\begin{footnotesize}
\vspace{-0.3cm}
\begin{align} \label{eq:rebar}
    \frac{\partial}{\partial \bphi} \E_{p_{\bphi}(\z)}[\Ld(\z)] &= 
    \underbrace{\E_{p_{\bphi}(\z)}\left[\left(\Ld(\z) - c_d(\z)\right) \partial_\bphi\!\log  p_{\bphi}(\z)\right]}_{\text{(i) reinforce}} \nonumber \\
    &\ - \underbrace{\E_{p_{\bphi}(\z)} \left[\partial_\bphi c_d(\z)  \right]}_{\text{(ii) correction}} + \underbrace{\partial_\bphi \E_{p_{\bphi}(\z)}[c_d(\z)]}_{\text{(iii) Gumbel-Softmax}}.%}
\end{align}
\end{footnotesize}
The gradient estimator in Eq.~\ref{eq:rebar} consists of three terms: (i) is the reinforce term, which is estimated using the Monte Carlo method by sampling $\z \sim p_\bphi(\z)$ and  $\bzeta \sim r_\bphi(\bzeta|\z)$. (ii) is the correction term due to the dependency of $c_d(\z)$ on $\bphi$. This term is approximated using the reparameterization trick applied to the conditional Gumbel-Softmax $r_\bphi(\bzeta|\z)$. (iii) is the Gumbel-Softmax term that can be written as

{\small
\vspace{-0.4cm}
\begin{equation} \label{eq:gsm}
\hspace{-0.3cm}
\E_{p_{\bphi}(\z)}[c_d(\z)] = \E_{p_{\bphi}(\z)}\left[ \E_{r_{\bphi}(\bzeta|\z)}[\Ld(\bzeta)] \right] = \E_{p_{\bphi}(\bzeta)} [\Ld(\bzeta)], 
\end{equation}
}%
which is the expected value of loss evaluated under the Gumbel-Softmax distribution $p_{\bphi}(\bzeta)$. Thus, its gradient can be computed also using the low-variance reparameterization trick. In practice, we only need two function evaluations for estimating the gradient in Eq.~\ref{eq:rebar}, one for computing $\Ld(\z)$, and one for $\Ld(\bzeta)$. The gradients are computed using an automatic differentiation library.

Eq.~\ref{eq:rebar} unifies the differentiable architecture search with policy gradient-based NAS methods~\cite{zoph2016neural, xie2018snas, tan2018mnasnet}. This estimator does not introduce any bias due to the continuous relaxation, as in expectation the gradient is equal to the REINFORCE estimator that operates on discrete variables. Moreover, this estimator uses the Gumbel-Softmax estimation of the differentiable loss for reducing the variance of the estimate. Under this framework, it is easy to see that SNAS~\cite{xie2018snas} is a biased estimation of the policy gradient as it only uses (iii) for search, ignoring other terms. On the other hand, policy gradient-based NAS~\cite{pham2018efficient,zoph2016neural, zoph2018learning} assumes a constant control variate ($c_d(\z) = C$) which only requires computing (i) as $\partial_\bphi \E_{p_{\bphi}(\z)}[C] = 0$.

{\bf Gradient Estimation for Non-Differentiable Loss  $\Ln$:} The gradient estimator in Eq.~\ref{eq:rebar} cannot be applied to non-differentiable loss $\Ln(\z)$ as the reparameterization trick is only applicable to differentiable functions. For $\Ln(\z)$, we use RELAX~\cite{grathwohl2017relax} that lifts this limitation by defining the baseline function $c_n(\z) := \E_{r_{\bphi}(\bzeta|\z)}[g(\bzeta)]$, where $g(.)$ is a surrogate function (\eg, a neural network) trained to be correlated with $\Ln(\z)$. The gradient estimator for $\Ln$ is obtained by replacing $c_d$ in Eq.~\ref{eq:rebar} with $c_n$:

\begin{footnotesize}
\vspace{-0.3cm}
\begin{align} \label{eq:relax}
    \frac{\partial}{\partial \bphi} \E_{p_{\bphi}(\z)}[\Ln(\z)] &= 
    \underbrace{\E_{p_{\bphi}(\z)}\left[\left(\Ln(\z) - c_n(\z)\right) \partial_\bphi\!\log  p_{\bphi}(\z)\right]}_{\text{(i) reinforce}} \nonumber \\
    &\ - \underbrace{\E_{p_{\bphi}(\z)} \left[\partial_\bphi c_n(\z)  \right]}_{\text{(ii) correction}} + 
    \underbrace{\partial_\bphi \E_{p_{\bphi}(\bzeta)}[g(\bzeta)]}_{\text{(iii) Gumbel-Softmax}},%}
\end{align}
\end{footnotesize}
However, the main difference is that here the reparameterization trick is applied to $\E_{r_{\bphi}(\bzeta|\z)}[g(\bzeta)]$ in (ii) and similarly to  $\E_{p_{\bphi}(\bzeta)}[g(\bzeta)]$ in (iii). Here, to make $g(\z)$ be correlated with $\Ln(\z)$, we train $g$ by minimizing $||g(\z) - \Ln(\z)||_2^2$. In the case of latency, this corresponds to training $g$ to predict latency on a set of randomly generated architectures before search. Similar to FBNet~\cite{wu2018fbnet} and ProxylessNAS~\cite{cai2018proxylessnas}, we use a simple linear function to represent $g(\z)$.

It is worth noting that the Gumbel-Softmax term, (iii) in Eq.~\ref{eq:relax}, minimizes the expectation of the approximation of the non-differentiable loss (e.g., latency) using the Gumbel-Softmax relaxation. This gradient estimator was used in FBNet~\cite{wu2018fbnet} for optimizing latency. In Eq.~\ref{eq:relax}, we can see that if $g$ cannot predict latency correctly, $\Ld(\z) - c_n(\z)$ will be large, thus, optimizing only (iii) will suffer from additional bias due to the approximation error. However, even if $g(\z)$ cannot approximate $\Ln(\z)$ accurately, for example in the case of compile-time performance optimizations, our gradient estimator is equal to the REINFORCE estimator, and it optimizes the true expected latency. Hence, UNAS does not suffer from any bias introduced due to the approximation of non-differentiable criteria.

\begin{figure*}
    \centering
    \vspace{-0.5cm}
    \includegraphics[width=0.75\textwidth]{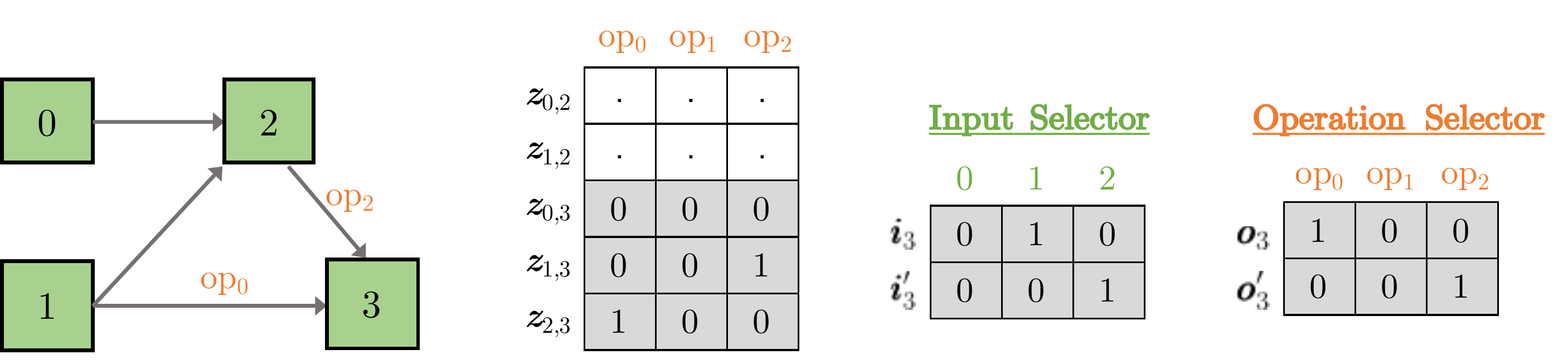}
    \caption{The factorized cell structure ensures that each node depends on two previous nodes. On the left, a small graph with 4 nodes is visualized. In the middle, $\z = \{\z_e\}$, the operation assignment for the incoming edges to node $3$ is shown. On the right, the input and operation selectors for these edges are shown. The shaded matrix on $\z$ is parameterized by the outer product $\i_{3} \otimes \o_{3} + \i'_{3} \otimes \o'_{3}$.}
    \label{fig:paired_input}
    % \vspace{-0.3cm}
\end{figure*}

\subsection{Training Objective}\label{sec:objective}
Several recent works on differentiable NAS have proposed bi-level training of architecture parameters and network parameters. In the architecture update, either training loss~\cite{xie2018snas}, or validation loss~\cite{liu2018darts} given the current network parameters $\w$, are used to update architecture parameters using
\begin{equation} \label{eq:phi_update}
    \min_{\bphi} \E_{p_{\bphi}(\z)}[\L_{\train}(\z, \w)], \ \text{or} \ \min_{\bphi} \E_{p_{\bphi}(\z)}[\L_{\val}(\z, \w)].
\end{equation}
Then, the network parameters $\w$ are updated given samples from the architecture by minimizing
\begin{equation} \label{eq:w_update}
    \min_{\w} \E_{p_{\bphi}(\z)}[\L_{\train}(\z, \w)].
\end{equation}
The parameters $\bphi$ and $\w$ are updated iteratively by taking a single gradient step in Eq.~\ref{eq:phi_update} and Eq.~\ref{eq:w_update}. It has been shown that by sharing network parameters among all the architecture instances, we gain several orders of magnitude speedup in search~\cite{liu2018darts,pham2018efficient}. However, this comes with the cost of updating architecture parameters at suboptimial $\w$. Intuitively, this translates to making decision on architecture without considering its optimal performance.

To avoid overfitting, we base our objective function on the generalization gap of an architecture. The rationale behind this is that the selected architecture not only should perform well on the training set, but also, should generalize equally well to the examples in the validation set, even if network weights are suboptimal. This prevents search from choosing architectures that do not generalize well. 
Formally, we define the generalization loss in search $\E_{p_{\bphi}(\z)}[\L_{\gen}(\z, \w)]$ by:
\begin{equation} \label{eq:phi_update_ge}
\hspace{-0.2cm}
    \E_{p_{\bphi}(\z)}\![\L_{\train}(\z, \w)\!+\!\lambda |\L_{\val}(\z, \w) - \L_{\train}(\z, \w)| ], 
\end{equation}
where $\lambda$ is a scalar balancing the training loss and generalization gap. We observe that $\lambda=0.5$ often works well in our experiments.\footnote{We also explored with the objective function without the absolute value, \ie, $\L_{\train}(\z, \w)\!+\!\lambda (\L_{\val}(\z, \w) - \L_{\train}(\z, \w))$. We observed that this variants does not perform as good as Eq.~\ref{eq:phi_update_ge}.} For training, we iterate between updating $\bphi$ using Eq.~\ref{eq:phi_update_ge} and updating $\w$ using Eq.~\ref{eq:w_update}. In each parameter update, we perform a simple gradient descent update.

{\bf Latency Loss:} In resource-constrained applications, we might be interested in finding an architecture that has a low latency as well as high accuracy. In this case, we can measure the latency of the network specified by $\z$ in each parameter update\footnote{We measure latency on the same hardware that the model is being trained.}. Representing the latency of the network using $\L_{lat}(\z)$, we augment the objective function in Eq.~\ref{eq:phi_update_ge} with $\E_{p_{\bphi}(\z)}[\lambda_{lat} \L_{lat}(\z)]$, where $\lambda_{lat}$ is a scalar balancing the trade-off between the architecture loss and the latency loss. Although $\L_{lat}(\z)$ is not differentiable \wrt $\z$, we construct a low-variance gradient estimator using Eq.~\ref{eq:relax} for optimizing this term.

%!TEX root = main.tex
\vspace{-0.2cm}
\section{Experiments in DARTS Search Space}
\label{sec:expts}
In this section, we apply the proposed UNAS framework to the problem of architecture search for image classification using DARTS~\cite{liu2018darts} search space, which was also used in~\cite{zoph2018learning, pham2018efficient, xie2018snas, chen2019pdarts}. We closely follow the experimental setup introduced DARTS~\cite{liu2018darts}. In the search phase, we search for a normal and reduction cell using a network with a small number of feature maps and/or layers. Given the stochastic representation of the architecture, the final cells are obtained by taking the configuration that has highest probability for each node as discussed below. Then, in the evaluation phase, the cells are stacked into a larger network which is retrained from scratch.
Sec.~\ref{sec:search} discuses a simple approach for factorizing cells that eliminates the necessity of post-search heuristics. Sec.~\ref{subsec:ablation} provides comparisons to previous work on three datasets.

\subsection{Factorized Cell Structure}
\label{sec:search}
Training the cell structure introduced in DARTS~\cite{liu2018darts} may result in a densely connected cell where each node depends on the output of all the previous nodes. In order to induce sparsity on the connectivity, prior work~\cite{chen2019pdarts,liu2018darts,zoph2018learning} heavily relies on post-search heuristics to limit the number of incoming edges for each node. % 
DARTS~\cite{liu2018darts} uses a heuristic to prune the number of input edges to two by choosing operations with the largest weights. P-DARTS~\cite{chen2019pdarts} uses an iterative optimization to limit the number of skip-connections and the number of incoming edges to two. The main issue with such post-search methods is that they create inconsistency between search and evaluation by constructing a cell structure without directly measuring its performance~\cite{xie2018snas}.

In order to explicitly induce sparsity, we factorize the operation assignment problem on the edges using two selection problems: i) an \textit{input selector} that selects two nodes out of the previous nodes and ii) an \textit{operation selector} that selects two operations that are applied to each selected input. We name this structure a \textit{factorized cell} as it enables us to ensure that the content of each node depends only on two previous nodes without relying on any post-search heuristic. Formally, we introduce $\i_{n}$ and $\i'_{n}$, two one-hot vectors for the n$^{th}$ node representing the input selectors as well as two one-hot vectors $\o_{n}$ and $\o'_{n}$ denoting the operation selectors. The architecture is specified by the sets $\{\i_{n}, \i'_{n}\}_{n=1}^{N}$ and $\{\o_{n}, \o'_{n}\}_{n=1}^{N}$, where $N$ is the number of nodes in a cell. 
This formulation is easily converted to the operation assignment problem on edges (\ie $\{\z_e\}$) in Sec.~\ref{sec:background} using the outer product $\i_{n} \otimes \o_{n} + \i'_{n} \otimes \o'_{n}$, as shown in Fig.~\ref{fig:paired_input}.
We use the product of categorical distributions in the form $\prod_n p(\i_{n}) p(\i'_{n}) p(\o_{n}) p(\o'_{n})$ to represent the distribution over architecture parameters. %In the following, we continue referring to this distributing by $p_{\bphi}(\z)$ for the ease of notation.

\subsection{Comparison with the Previous Work}\label{subsec:ablation}
The current literature on NAS often reports the final performance obtained by the best discovered cell. Unfortunately, such qualitative metric fails to capture i) the number of searches conducted before finding the best cell, ii) the performance variation resulted from different searches, iii) the effect of each model component on the final performance, and iv) the effect of post-search heuristics used for creating the best architecture. To better provide insights into our framework, we conduct extensive ablation experiments on the CIFAR-10, CIFAR-100 and ImageNet datasets. We run the search and evaluation phases end-to-end four times on each dataset and we report mean and standard deviation of the final test error as well as the best cell out of the four searches. We do not use any post-search heuristic, as our factorized cell structure always yields two-incoming edges per node in the cell. This stands in a stark contrast to DARTS~\cite{liu2018darts} and P-DARTS~\cite{chen2019pdarts} that use post-search heuristics to sparsify the discovered cell.

Here, we only consider the differentiable cross-entropy loss functions as the search objective function (\ie, we do not optimize for latency). Since the direct search on ImageNet is computationally expensive, we reduce the search space on this dataset to five operations including skip connection, depthwise-separable 3$\times$3 convolution, max pooling, dilated  depthwise-separable 3$\times$3 convolution, and depthwise-separable 5$\times$5 convolution. Prior work on ResNets~\cite{he2016deep}, DenseNets~\cite{huang2017densely}, as well as the recent RandWire~\cite{xie2019exploring} suggest that it should be possible to achieve high accuracy by using only these three operations.

Below, we discuss the different baselines summarized in Table~\ref{table:cifar10_ab}. Additional details of search and evaluation can be found in Appendix~\ref{app:search_settings}, and  Appendix~\ref{app:eval_settings} respectively.

\begin{table*}
\caption{Comparison against the state-of-the art methods. Different objective functions for updating architecture parameters and different gradient estimators are examined for UNAS. We run UNAS and the original publicly-available source code for DARTS~\cite{liu2018darts} and P-DARTS~\cite{chen2019pdarts} end-to-end four times with different initialization seeds. Mean$\pm$standard deviation of all four discovered architectures as well as the best architecture at the end of the evaluation phase are reported. For other techniques, the original best results are reported. The search cost is reported on CIFAR-10.
UNAS with $\L_{\gen}$ and REBAR significantly outperforms gradient-based methods on all three datasets.} \label{table:cifar10_ab}
\centering
%\resizebox{1.0\linewidth}{!}{
%\setlength{\tabcolsep}{2pt}
    \begin{tabular}{cccccccccc}
        \toprule
        & \multicolumn{1}{c}{\bf Objective}     & \bf  Gradient&  \multicolumn{2}{c}{\bf CIFAR-10} &  \multicolumn{2}{c}{\bf CIFAR-100}     &  \multicolumn{2}{c}{\bf ImageNet} & \bf Search Cost \\
        & \multicolumn{1}{c}{\bf Function}      & \bf Estimator & mean  & best & mean & best & mean & best & (GPU days) \\
        \midrule
        \multirow{3}{*}{\rotatebox{90}{\small \bf UNAS}} & $\L_{\val}$ &  Gumbel-Soft. & 2.79{\tiny$\pm$0.10} & 2.68 & 17.11{\tiny$\pm$0.38} & 16.80 & 26.06{\tiny$\pm$0.51} & 25.41 & - \\
        & $\L_{\gen}$ &  Gumbel-Soft. & 2.81{\tiny$\pm$0.01} & 2.74 & 16.98{\tiny$\pm$0.34} & 16.59 & 24.64{\tiny$\pm$0.13} & \bf 24.46 & - \\
        %\midrule
        & $\L_{\gen}$ & REBAR         & \bf 2.65{\tiny$\pm$0.07} & \bf 2.53 & \bf 16.72{\tiny$\pm$0.76} & \bf 15.79 & \bf 24.60{\tiny$\pm$0.06} & 24.49 & 4.3 \\
        %$\L_{\gen}$ & REINFORCE     &    &   & - \\
        \midrule
        \multirow{3}{*}{\rotatebox{90}{\small \bf Gradient}} & \multicolumn{2}{l}{DARTS~\cite{liu2018darts}}   &  3.03{\tiny$\pm$0.16} & 2.80 & 27.83{\tiny$\pm$8.47} & 20.49 & 25.27{\tiny$\pm$0.06} & 25.20 & 4 \\
        & \multicolumn{2}{l}{P-DARTS~\cite{chen2019pdarts}} & 2.91{\tiny$\pm$0.14} & 2.75 & 18.09{\tiny$\pm$0.49} & 17.36 & 24.98{\tiny$\pm$0.44} & 24.49 & 0.3 \\
        & \multicolumn{2}{l}{SNAS~\cite{xie2018snas}} & - & 2.85 & - & - & - & 27.3 & 1.5 \\
        \midrule
        \midrule
        \multirow{3}{*}{\rotatebox{90}{\small \bf Reinforce}} & \multicolumn{2}{l}{NASNet-A ~\cite{zoph2018learning}}    & - & 2.65 & - & - & - & 26.0 & 2000 \\
        & \multicolumn{2}{l}{BlockQNN~\cite{zhong2018practical}} & - & 3.54 & - & 18.06 & - & - & 96 \\
        & \multicolumn{2}{l}{ENAS~\cite{pham2018efficient}} & - & 2.89 & - & - & - & - & 0.45\\
        \midrule
        \multirow{3}{*}{\parbox[t][0.6cm]{2mm}{\multirow{3}{*}{\rotatebox[origin=c]{90}{\small \bf Evolution}}}}& \multicolumn{2}{l}{AmoebaNet-A~\cite{real2018regularized}} & - & 3.12 & - & - & - & 25.5 & 3150 \\
        & \multicolumn{2}{l}{AmoebaNet-B~\cite{real2018regularized}} & - & 2.55 & - & - & - & 26.0 & 3150\\
        & \multicolumn{2}{l}{AmoebaNet-C~\cite{real2018regularized}} & - & - & - & - & - & 24.3 & 3150 \\
        & \multicolumn{2}{l}{Hierarchical. Evolution~\cite{liu2017hierarchical}} & - & 3.75 & - & - & - & - & 300\\
        \bottomrule
    \end{tabular}%
%}
\end{table*}

\begin{figure*}
\begin{minipage}[t]{.59\textwidth}
\centering
\begin{subfigure}[b]{.4\textwidth}
    \setlength{\belowcaptionskip}{0pt}
    %\vspace{-0.2cm}
    \vspace{-0.4cm}
    \includegraphics[trim={1cm 0 0 0},width=1.05\linewidth]{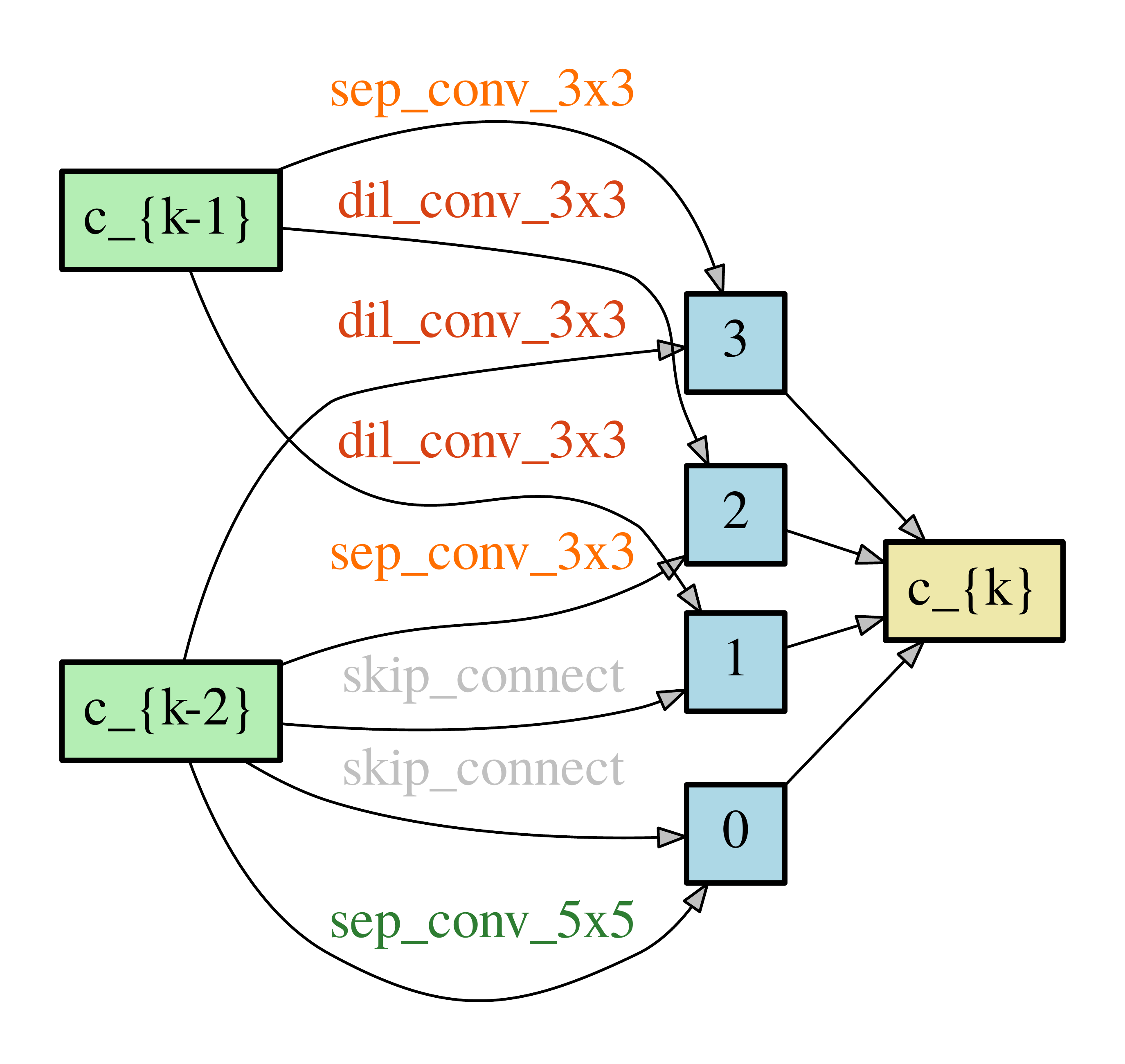}
    \caption{Normal Cell}
    \label{fig:cifar10_normal}
\end{subfigure}%
\begin{subfigure}[b]{.65\textwidth}
    \setlength{\belowcaptionskip}{0pt}
    %\vspace{-0.2cm}
    \vspace{-0.4cm}
    \includegraphics[trim={0cm 0 0 0cm},width=1.05\linewidth]{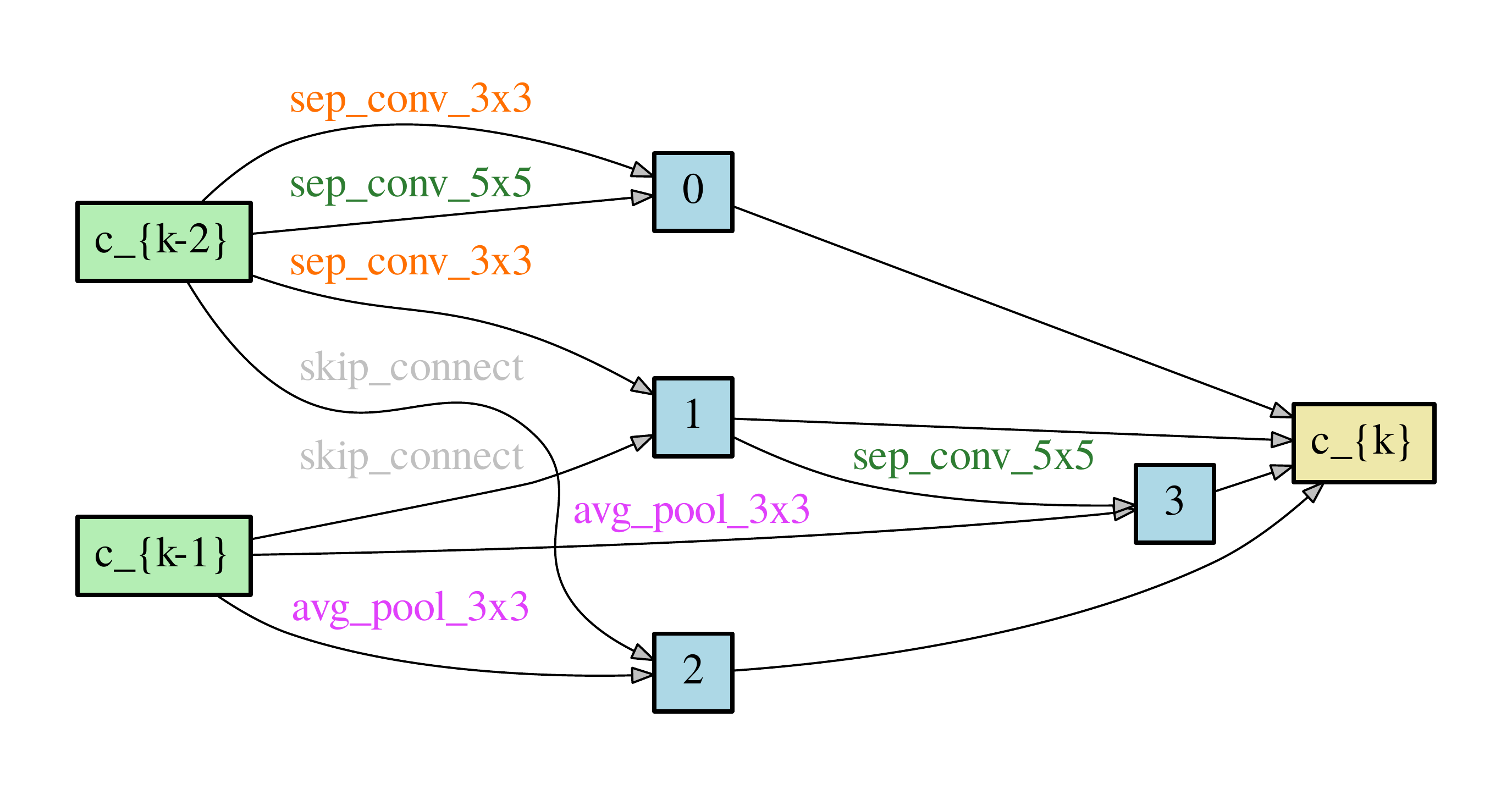}
    \caption{Reduce Cell}
    \label{fig:cifar10_reduce}
\end{subfigure}
\caption{The best performing cell on CIFAR-10.}
\label{fig:cifar10_best_cell}
\end{minipage} \hfill
\begin{minipage}[t]{.39\textwidth}
        \centering
        \begin{subfigure}[t]{1\textwidth}
        \centering
            % \vspace{-2cm}
             \includegraphics[width=0.9\linewidth]{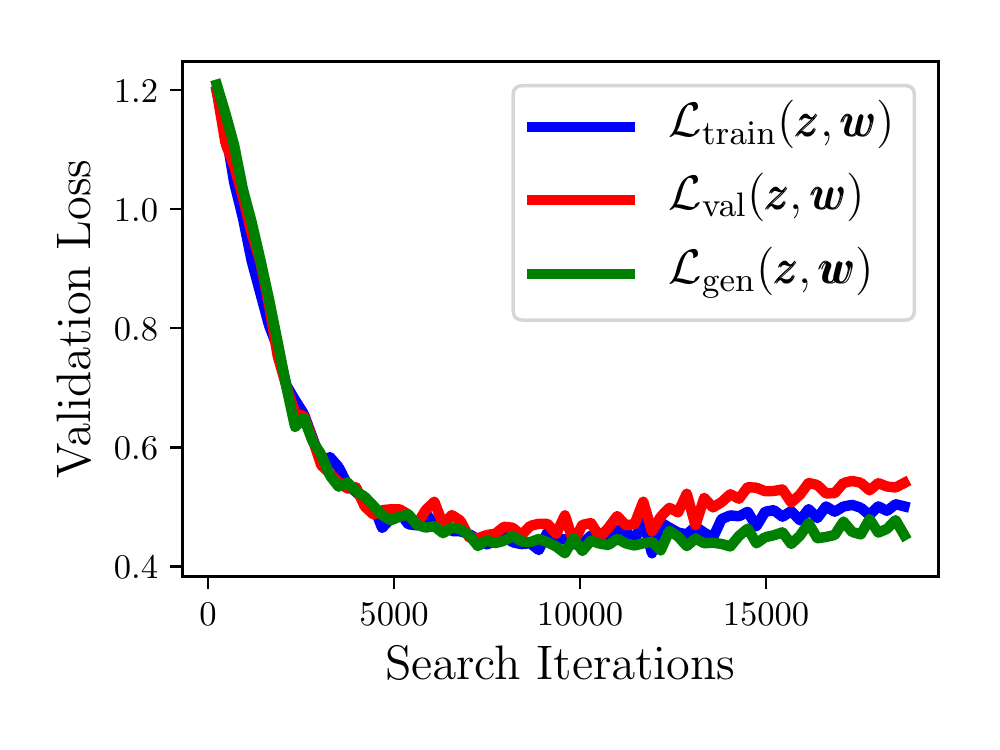}
             \vspace{-0.8cm}
        \end{subfigure}
    \caption{Validation loss in search.}
    \label{fig:objective_fun}
\end{minipage} 
\end{figure*}

\textbf{The state-of-the-art}: The previous works closest to our work DARTS~\cite{liu2018darts}, P-DARTS~\cite{chen2019pdarts} and SNAS~\cite{xie2018snas} have unfortunately reported the performance for the best discovered cell. Since DARTS and P-DARTS implementations are publicly available, for a fair comparison, we run their original source code end-to-end four times similar to our model with different random initialization seeds using hyperparameters and commands released by the authors on CIFAR-10 and CIFAR-100.\footnote{We exactly followed the hyperparameters and commands using the search/eval code provided by the authors. We only set the initialization seed to a number in $\{0, 1, 2, 3\}$.} For the ImageNet datasets, we transfer the discovered cells from CIFAR-10 to this dataset as described in~\cite{liu2018darts, chen2019pdarts}.  The implementation of SNAS~\cite{xie2018snas} is not publicly available. So, we compare against this work using the original published results. Finally, in order to better contextualize our results, we compare UNAS against previous methods that use reinforcement learning or evolutionary search. On ImageNet, we only consider the mobile-setting (FLOPS $<$ 600M) which is often used to compare NAS models.

\textbf{UNAS baselines:} We also explore the different variants of UNAS. The objective function column in Table~\ref{table:cifar10_ab} represents the loss function used during search for updating $\bphi$. Here, $\L_{\val}$ (Eq.~\ref{eq:phi_update}) and $\L_{\gen}$ (Eq.~\ref{eq:phi_update_ge}) are considered. The gradient estimator column represents the gradient estimator used for updating $\bphi$ during search. We examine Gumbel-Softmax and REBAR (Eq.~\ref{eq:rebar}).

\begin{figure*}[h]
\centering
\vspace{-0.5cm}
\begin{subfigure}{.9\textwidth}\centering
    \includegraphics[scale=0.4, trim={2cm 4.cm 2cm 10cm}, clip=True]{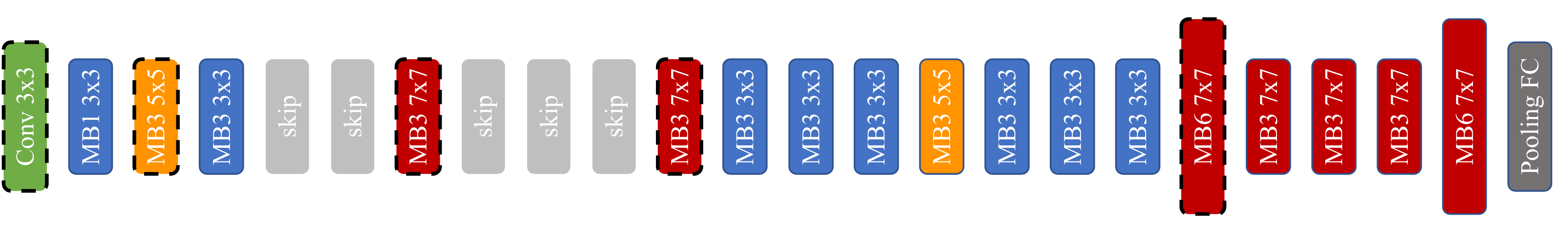}
    \caption{Cell discovered by UNAS in the ProxylessNAS~\cite{cai2018proxylessnas} search space with 9.8 ms GPU latency and 24.7\% top-1 error}
\end{subfigure} \\
\begin{subfigure}{.9\textwidth}\centering
    \includegraphics[scale=0.4, trim={2cm 12cm 2cm 2cm}, clip=True]{images/unas_proxyless.pdf}
    \caption{Cell discovered by ProxylessNAS~\cite{cai2018proxylessnas} with 10.1 ms GPU latency and  24.9\% top-1 error}
\end{subfigure}
\caption{Visualization of the network discovered by UNAS in the ProxylessNAS~\cite{cai2018proxylessnas} search space. MB$e\ K\!\times\!K$ denotes a mobile inverted residual block with expansion ratio $e$ and kernel size $K$. UNAS, in contrast to ProxylessNAS, keeps the cells at the deeper layers (on the right side) computationally inexpensive by using a small expansion ratio, enabling more MBConv layers in the shallower layers. Although UNAS architecture is deeper, it has a lower latency with the same network width.}
\label{fig:compare_proxless}
\end{figure*}

\textbf{Observations:} From the first group of Table~\ref{table:cifar10_ab},
we observe that architecture search with the generalization loss yields a better model often in terms of both average performance and best results. The improvement obtained by the generalization is especially profound in ImageNet as this loss function improves $\L_{\val}$ by 1.4\% in average. We can also see that our REBAR gradient estimator often improves the results across all datasets. From the second group of Table~\ref{table:cifar10_ab}, we observe that our UNAS framework with REBAR estimator and the generalization loss significantly outperforms DARTS, P-DARTS, and SNAS on all three datasets.\footnote{The significance test between UNAS and any other approach passes on all the datasets with p-value $< 0.05$, except on ImageNet between UNAS and P-DARTS which yields p-value $=0.18$.} Interestingly, our full model ($\L_{\gen}$ with REBAR) exhibits a low variance on CIFAR-10 and ImageNet, showing the robustness of the framework in discovering high-performing architectures.
Finally, comparing UNAS against the evolutionary and RL-based models shows that UNAS outperforms these models. The only exception is AmoebaNet-C~\cite{real2018regularized} on ImageNet. However, note that this method requires 700x more GPU time to search.

\textbf{Why does the generalization loss help in search:} Recall that in differentiable architecture search, we often update the architecture distribution parameters $\bphi$ using suboptimal $\w$. We hypothesize that even if validation loss is used in search due to the suboptimality of $\w$, the architecture is not discovered using the the true generalization of the network to unseen examples. To illustrate this, the validation loss during architecture search visualized in Fig.~\ref{fig:objective_fun} for different loss functions. We observe that even using the validation loss for updating architecture parameters does not prevent the network from overfitting.

One question is whether our generalization loss is required in the case of the original RL-based NAS~\cite{pham2018efficient, zoph2016neural, zoph2018learning}, which updates architecture parameters using $\w$ closer to optimality. To answer this, we also examine with ENAS~\cite{pham2018efficient}-like training where network parameters $\w$ are updated for half epoch in every $\bphi$ update (\ie, the network parameters $\w$ are brought closer to the optimum). In this case, the architectures found by generalization loss in average obtains test error 2.92\% on CIFAR-10 compared with the validation loss based search that achieves 3.12\%. This provides another evidence that architecture search can potentially benefit from considering generalization, opening up new research directions in NAS.

\textbf{Cell visualization:} The best cell discovered on the CIFAR-10 dataset is visualized in Fig.~\ref{fig:cifar10_best_cell}. See Appendix~\ref{app:cell_vis} for the visualization of best cells on other datasets.

\textbf{More comparisons:} We provide in-depth comparisons against the state-of-the-art techniques with more detailed information including the number of parameters, search cost, and the number of floating point operations in Appendix.~\ref{app:previous}.

\begin{figure}[h!]
    \centering
        \begin{minipage}[t]{.5\textwidth}
        \centering
        \setlength{\tabcolsep}{1pt}
        \captionof{table}{Latency-based architecture search. Models are sorted based on their top-1 error. For a better illustration, Fig.~\ref{fig:latency} compares the models visually.}
        \label{table:latency}
        %\vspace{0.25cm}
        %\resizebox{\linewidth}{!}{%
            \begin{tabular}{lccc}
                \toprule
                \multirow{2}{*}{\bf Architecture} & \multicolumn{2}{c}{\bf \underline{\ Val Error\ }} & {\bf Latency} \\
                & top-1 & top-5 & (ms) \\
                \midrule
                %RandWire-WS~\cite{xie2019exploring} & 25.3 & 7.8 & & 5.6 & 583 \\
                % MnasNet-A1~\cite{tan2018mnasnet}    & 24.8 & 7.5 & 3.9 & 312 \\
                % MnasNet-A2~\cite{tan2018mnasnet}    & 24.4 & 7.3 & 4.8 & 340 \\
                EfficientNet B0~\cite{tan2019efficientnet}                     & 23.7  & 6.8  & 14.5 \\
                MobileNetV3 Large~\cite{Howard2019Mobilenetv3} & 24.7  & 7.6  & 11.0 \\
                MnasNet-A1~\cite{tan2018mnasnet}    & 24.8  & 7.5  & 10.9 \\
                Single-Path NAS~\cite{stamoulis2019singlepath}             & 25.0  & 7.8  & 10.2 \\
                FBNet-C~\cite{wu2018fbnet}          & 25.1  & -    & 11.5 \\
                MobileNetV2 1.4x~\cite{sandler2018mobilenetv2} & 25.3  & 7.5 & 13.0 \\
                MnasNet-B1~\cite{tan2018mnasnet}    & 25.5 & - &  9.4 \\
                ShuffleNet V2 2x~\cite{ma2018shufflenet} & 26.3 & - & 9.16 \\
                MobileNetV2 1x~\cite{sandler2018mobilenetv2} & 28.0  & 9. & 9.2 \\
                \midrule
                ProxylessNAS-GPU~\cite{cai2018proxylessnas} & 24.9 & 7.5 & 10.1 \\
                \midrule
                UNAS                          & 24.7 & 7.6 & 9.8 \\
                \bottomrule
            \end{tabular}%
        %}
    \end{minipage}
    \begin{minipage}[t]{.45\textwidth}
        \begin{subfigure}{0.95\textwidth}
            %\setlength{\belowcaptionskip}{-10pt}
            %\vspace{-0.2cm}
            \centering
            \includegraphics[width=\linewidth]{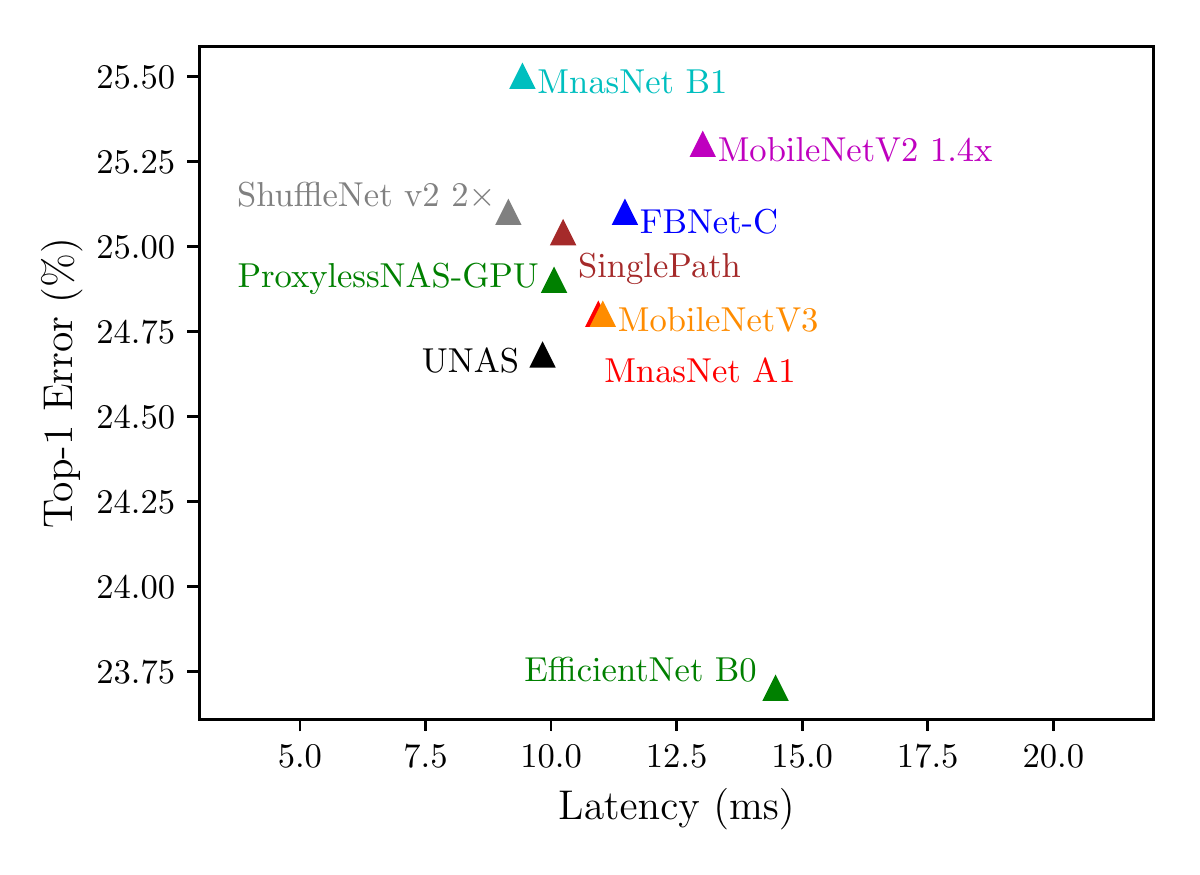}
            % \caption{Accuracy vs. Latency}
            % \label{fig:imagenet_reduce}
        \end{subfigure}
        \vspace{-0.3cm}        
        \caption{Latency-based architecture search. We seek architectures that are in the bottom-left side of the error vs. latency axes. UNAS discovers an architecture that is more accurate and has a low latency compared to the current state-of-the-art architectures based on MobileNetV2.}
        \label{fig:latency}
    \end{minipage}
    \hfill
%\end{figure}
%\begin{figure}
%    % \setlength{\belowcaptionskip}{-10pt}
%    \centering 
\iffalse
    \begin{minipage}[t]{.47\textwidth}
        \begin{subfigure}{.5\textwidth}
            \setlength{\belowcaptionskip}{-5pt}
            %\vspace{-0.3cm}
            \includegraphics[trim={1cm 0 0 0},width=1.05\linewidth]{images/imagenet_normal.pdf}
            % \vspace{-0.3cm}
            \caption{Normal Cell}
            \label{fig:imagenet_normal}
        \end{subfigure}%
        \begin{subfigure}{.5\textwidth}
            \setlength{\belowcaptionskip}{-5pt}
            %\vspace{-0.3cm}
            \includegraphics[trim={1cm 0 0 0},width=1.05\linewidth]{images/imagenet_reduce.pdf}
            % \vspace{-0.3cm}
            \caption{Reduce Cell}
            \label{fig:imagenet_reduce}
        \end{subfigure}
        \caption{The best cell on ImageNet.}
        \label{fig:best_imagenet_cell}
    \end{minipage}
\fi
\end{figure}

\section{Experiments on Latency-based Search} \label{subsec:latency}
In this section, we examine our proposed framework for searching architectures with low latency directly on the ImageNet dataset. Unfortunately, the DARTS search space results in high-latency networks due to the parallel branches and concatenation in each cell. So, here, we change the building blocks of our search space to the mobile inverted bottleneck convolution (MBConv)~\cite{sandler2018mobilenetv2} that has been used in ProxylessNAS~\cite{cai2018proxylessnas} and FBNet~\cite{wu2018fbnet} for discovering low-latency networks. For this section, we closely follow the search space introduced in ProxylessNAS~\cite{cai2018proxylessnas} for ImageNet in which a 21-layer network with seven choices of operations in each layer is searched. Specifically, for each layer, an MBConv is selected among various kernel sizes $\{3, 5, 7\}$ and expansion ratios $\{3, 6\}$. To allow layer removal, an additional skip-connection is used in ProxylessNAS yielding seven operations per layer. For search and evaluation we closely follow the settings used in ProxylessNAS (see Appendix~\ref{app:latency_settings} for details).

For gradient estimation of the latency loss in UNAS, we use Eq.~\ref{eq:relax} with a simple linear function as the surrogate function, \ie, $g(\z) = \sum_{i,j} l_{i, j} z_{i,j}$ where $z_{i, j} \in \{0, 1\}$ is a binary scalar indicating if operation $i$ is used in layer $j$ and $l_{i, j}$ is the approximate latency associated with the operation. Similar to ProxylessNAS, we randomly generate 10K network samples before search and we train the parameters of $g$ (\ie, all $l_{i, j}$) by minimizing an $L_2$ regression loss.

We search for architecture on V100 GPUs, as it allows us to measure the true latency on the device during search. These GPUs were also used in ProxylessNAS~\cite{cai2018proxylessnas} which enables us to have a fair comparison against this method. We measure latency using a batch size of 32 images. We empirically observed that smaller batch sizes under-utilize GPUs, resulting in inaccurate latency measurements.

Table~\ref{table:latency} and Fig.~\ref{fig:latency} report the latency and validation set error on ImageNet for our model in comparison to recent hardware-aware NAS frameworks that operate in a similar search space (\ie, MobileNetV2~\cite{sandler2018mobilenetv2}) and have similar latency ($\sim$10 ms on V100 GPUs). We can see that UNAS finds an architecture that is slightly faster but more accurate than the ProxylessNAS-GPU~\cite{cai2018proxylessnas} architecture that uses exactly the same search space and the same target device. EfficientNet B0~\cite{tan2019efficientnet} is the only architecture that is more accurate than UNAS but it is also 48\% slower on the GPU. Although EfficientNet B0 has a low number of mathematical operations, it is not so efficient on TPU/GPU due to the heavy usage of depth-wise separable convolutions~\cite{efficientnet_edge_tpu}.
The architectures that are faster than UNAS including ShuffleNet v2~\cite{ma2018shufflenet}, MnasNet B1~\cite{tan2018mnasnet} and MobileNetV2 1.0x~\cite{sandler2018mobilenetv2} are also less accurate.\footnote{All models are examined in PyTorch. For MobileNetV2 and ShuffleNet V2 the official PyTorch implementations are used. For ProxylessNAS-GPU, the original code provided in \cite{cai2018proxylessnas} is used. Other networks implementations are obtained from EfficientNets repo~\cite{wightman_2019} which are optimized for PyTorch.} 

In Fig.~\ref{fig:compare_proxless}, the architecture discovered by UNAS is compared against ProxylessNAS-GPU that has been discovered for the same type of GPUs. Interestingly, UNAS discovers an architecture that is deeper, \ie, it has 3 more MBConv layers. But, it also
faster and more accurate than the architecture discovered by ProxylessNAS.

%!TEX root = main.tex
\vspace{-0.2cm}
\section{Conclusions}
% NAS aims for automating the manual and costly process of architecture design. 
In this paper, we presented UNAS that unifies differentiable and RL-based NAS. Our proposed framework uses the gradient of the objective function for search without introducing any bias due to continuous relaxation. In contrast to previous DNAS methods, UNAS search objective is not limited to differentiable loss functions as it can also search using non-differentiable loss functions. We also introduced a new objective function for search based on the generalization gap and we showed that it outperforms previously proposed training or validation loss functions. 

In extensive experiments in both DARTS~\cite{liu2018darts} and ProxylessNAS~\cite{cai2018proxylessnas} search spaces, we showed that UNAS finds architectures that 1) are more accurate on CIFAR-10, CIFAR-100, and ImageNet and 2) are more efficient to run on GPUs. We will make our implementation publicly available to facilitate the research in this area.

{\small
\bibliographystyle{ieee_fullname}
\bibliography{egbib}
}

\appendix
\clearpage
%!TEX root = main.tex

\appendix
\section*{\Large Appendix}

\section{Architecture Search Settings}\label{app:search_settings}
In this section, the implementation details for the search phase are provided.

\subsection{Search Space}
We use the following 7 operations in our search on CIFAR-10 and CIFAR-100:
\begin{enumerate}[noitemsep,topsep=0pt]
    \item \verb|skip_connect|: identity connection
    \item \verb|sep_conv_3x3|: depthwise-separable 3x3 convolution
    \item \verb|max_pool_3x3|: max pooling with 3x3 kernel
    \item \verb|dil_conv_3x3|: dilated depthwise-separable 3x3 convolution
    \item \verb|sep_conv_5x5|: depthwise-separable 5x5 convolution
    \item \verb|avg_pool_3x3|: average pooling with 3x3 kernel
    \item \verb|sep_conv_7x7|: depthwise-separable 7x7 convolution
\end{enumerate}
In the case of ImageNet, in order to make the search tractable, we only use the first five operations. All operations use a stride of 1 when part of the Normal Cell, and a stride of 2 when part of the Reduce Cell. Appropriate padding is added to the input features to preserve the spatial dimensions. Each convolution consists of a (\texttt{ReLU-Conv-BN}) block, and the depthwise separable convolutions are always applied twice, consistent with prior work~\cite{liu2018darts,real2018regularized,xie2018snas,zoph2018learning}.

\subsection{CIFAR-10 and CIFAR-100} \label{app:cifar10_search}
The CIFAR-10 and CIFAR-100 datasets consist of 50,000 training images and 10,000 test images.
During search, we use 45,000 images from the original training set as our training set and the remaining as the validation set. The final evaluation phase uses the original split. 
During architecture search, a network is constructed by stacking 8 cells with 4 hidden nodes. Similar to DARTS~\cite{liu2018darts}, the cells are stacked in the blocks of 2-2-2 Normal cells with Reduction cells in between. The networks are trained using 4 Tesla V100 GPUs with a batch size of 124, for 100 epochs. For the first 15 epochs, only the network parameters ($\w$) are trained, while the architecture parameters ($\bphi$) are frozen. This pretraining phase prevents the search from ignoring the operations that are typically slower to train. The architecture parameters are trained using the Adam optimizer with cosine learning rate schedule starting from 2$\times10^{-3}$ annealed down to 3$\times10^{-4}$. The network parameters are also trained using Adam with cosine learning rate schedule starting from 6$\times10^{-4}$ annealed down to 1$\times10^{-4}$. We use 
$\lambda=0.5$, and a Gumbel-Softmax temperature of 0.4.

One issue with the factorized structure is that the architecture search may choose the same input and operation pair for both incoming edges of a node due to the symmetric expression in $\i_{n} \otimes \o_{n} + \i'_{n} \otimes \o'_{n}$. To prevent this, we add an architecture penalty term to our objective function using $\L_{arch}(\z) = \E \left[\lambda_{arch}\sum_{n=1}^{N} \text{tr}([\i_{n} \otimes \o_{n}] [\i'_n \otimes \o'_{n}]^T)\right]$ where $\lambda_{arch}$ is a trade-off parameter ($\lambda_{arch}=0.2$). The term inside the summation is one if the same input/op pairs are selected by $(\i_{n}, \o_{n})$ and $(\i'_{n}, \o'_{n})$.

\subsection{ImageNet}
\label{app:imagenet_search}
We search using a 14-layer network with 16 initial channels, over 8 V100 GPUs, needing around 2 days. We use a learning rate of 3$\times10^{-4}$ with Adam to learn the network parameters of the mixed-op network. We train architecture parameters with a learning rate of 1$\times10^{-3}$ using Adam. We parallelize training over 8 GPUs without scaling the learning rate. For the first 5 epochs, we only train the network parameters ($\w$), and in the remaining 15 epochs, we update both $\w$ and $\bphi$. We use 
$\lambda=0.5$ and $\lambda_{arch}=0.2$, the same as CIFAR-10, and a Gumbel-Softmax temperature of 0.4. We use a weight decay of 3$\times10^{-4}$ on the weight parameters, and 1$\times10^{-6}$ on the architecture parameters. 90\% of the ImageNet train set is used to train the weight parameters, while the rest is used as the validation set for training the architecture parameters.

\section{Architecture Evaluation Settings}\label{app:eval_settings}
In this section, the implementation details for the evaluation phase are provided.

\subsection{CIFAR-10 and CIFAR-100}\label{app:cifar10_eval}
% \textbf{Evaluation:} 
The final network is constructed by stacking a total of 20 cells. The networks are trained on a V100 GPU with a batch size of 128 for 600 epochs. SGD  with momentum 0.9 is used. The cosine learning rate schedule is used starting from 5$\times10^{-2}$ annealed down to zero. Similar to DARTS, the path dropout of the probability 0.2 on CIFAR-10 and 0.3 on CIFAR-100, and cutout of 16 pixels are used.

\subsection{ImageNet}
\label{app:imagenet_eval}
% {\bf Data Augmentation.} 
For data augmentation, we use the same settings as DARTS~\cite{liu2018darts}. We randomly crop training images to a size of 224$\times$224 px along with a random horizontal flip, and jitter the color. During evaluation, we use a single center crop of size 224$\times$224 px after resizing the image to 256$\times$256 px.

% {\bf Hyperparameters.}
For the final evaluation, we train a 14 layer network for 250 epochs with an initial channel count such that the multiply-adds of the network is $<$600M, as per the mobile setting proposed by~\cite{howard2017mobilenets}. We train our networks using SGD with momentum of 0.9, base learning rate of 0.1, weight decay of 3$\times10^{-5}$, with a batch size of 128 per GPU. We train our model for 250 epochs in line with prior work~\cite{liu2018darts,xie2019exploring,xie2018snas}, annealing the learning rate to 0 throughout the training using a cosine learning rate decay. We scale training to 8 V100 GPUs using the linear scaling rule proposed in~\cite{goyal2017accurate}, with a learning rate warmup for the first 5 epochs.

\iffalse
\section{Errors obtained on training the best models multiple times.}
Table~\ref{table:multiple_best_results_imagenet} shows the results obtained when only the best models are trained 5 times, with different random seeds. The lowest errors we obtained over 5 runs were 24.74\% and 7.63\% for top-1 and top-5 errors respectively, using the cell searched on ImageNet.
\begin{table}[h]
    \caption{ImageNet performance of the best models, averaged over 5 evaluation runs.}
    \label{table:multiple_best_results_imagenet}
    \centering
    \begin{tabular}{lcccc}
        \toprule
        \multirow{2}{*}{\bf Architecture} & \multicolumn{2}{c}{\bf \underline{Val Error (\%)}} & {\bf Params} & {\bf MA}\\
        & top-1 & top-5 & (M) & (M) \\
                \midrule
        \multicolumn{5}{c}{\bf Random best chosen from 10 models, averaged over 5 eval runs}\\
        Random Best & 25.61$_{\pm 0.07}$ & 8.09$_{\pm 0.10}$ & 5.37 & 597.61 \\
        \midrule
        \multicolumn{5}{c}{\bf Best cell searched on CIFAR-10, averaged over 5 eval runs}\\
        Searched on CIFAR-10 Best & 25.38$_{\pm 0.17}$ & 7.88$_{\pm 0.08}$ & 5.13 & 574.61 \\
        \midrule
        \multicolumn{5}{c}{\bf Best cell searched on ImageNet, averaged over 5 eval runs}\\
        Searched on ImageNet Best  & 25.02$_{\pm 0.36}$ & 7.73$_{\pm 0.20}$ & 5.39 & 599.69 \\
        \bottomrule
    \end{tabular}
\end{table}
\fi

\section{Best Cell Structures}
\label{app:cell_vis}

% \subsection{Best Cell discovered on ImageNet}
\begin{figure}[h]
    \setlength{\belowcaptionskip}{0pt}
    \centering
    \caption{The best performing cell discovered on ImageNet.}
    \label{fig:ap_imagenet_best_cell}
    \begin{subfigure}{.5\textwidth}
    \centering
        \caption{Normal Cell}
        \label{fig:ap_imagenet_normal}
        \includegraphics[trim={0 0 0 1.5cm},clip,scale=0.15]{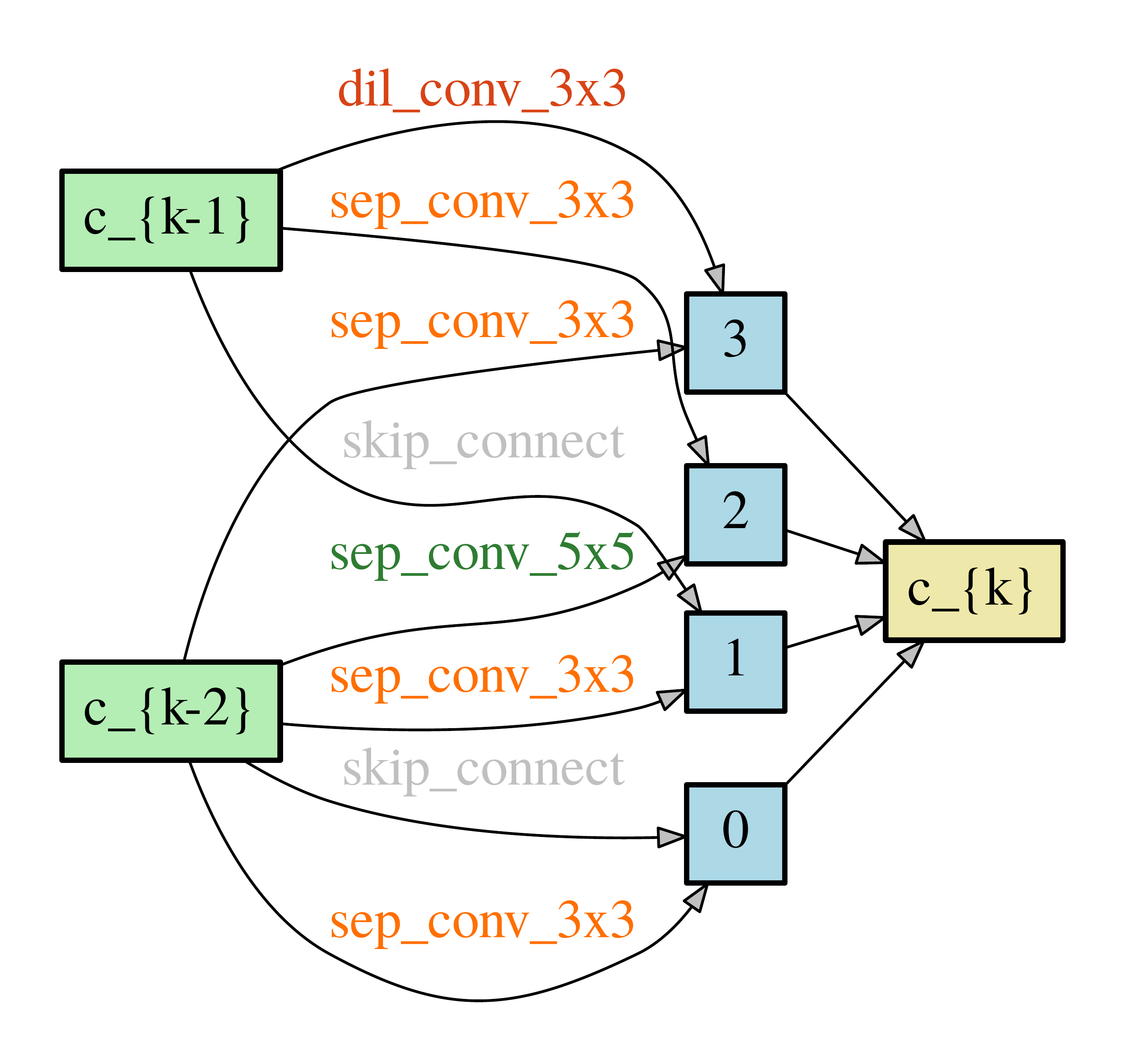}
    \end{subfigure}
    \begin{subfigure}{.5\textwidth}
    \centering
        \caption{Reduce Cell}
        \label{fig:ap_imagenet_reduce}
        \includegraphics[trim={0 0 0 1.5cm},clip,scale=0.25]{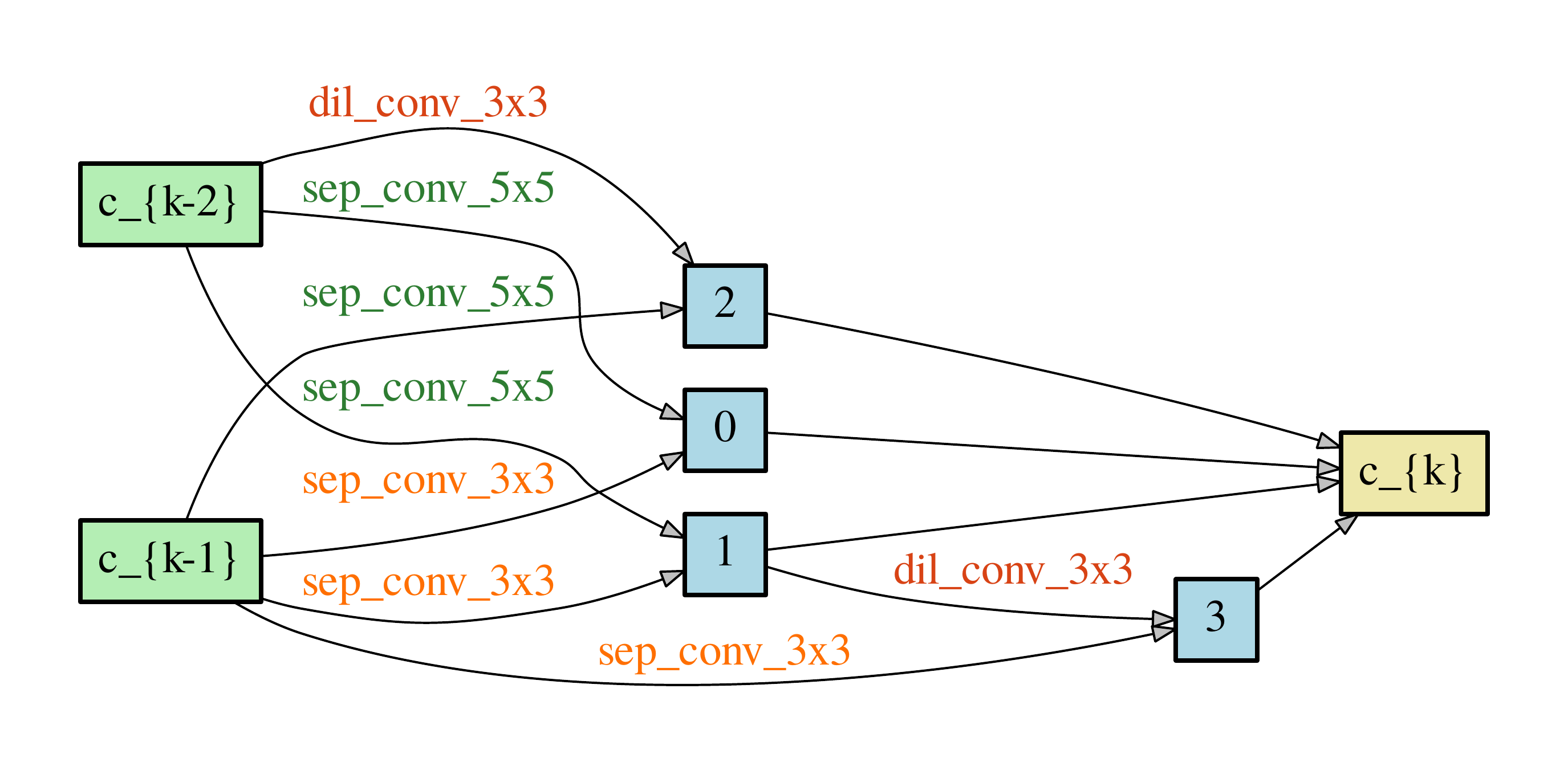}
    \end{subfigure}
\end{figure}

% \subsection{Best Cell discovered on CIFAR-10}
\begin{figure}[h]
    \setlength{\belowcaptionskip}{0pt}
    \centering
    \caption{The best performing cell discovered on CIFAR-10.}
    \label{fig:ap_cifar10_best_cell}
    \begin{subfigure}{.3\textwidth}
    \centering
        \caption{Normal Cell}
        \label{fig:ap_cifar10_normal}
        \includegraphics[scale=0.15]{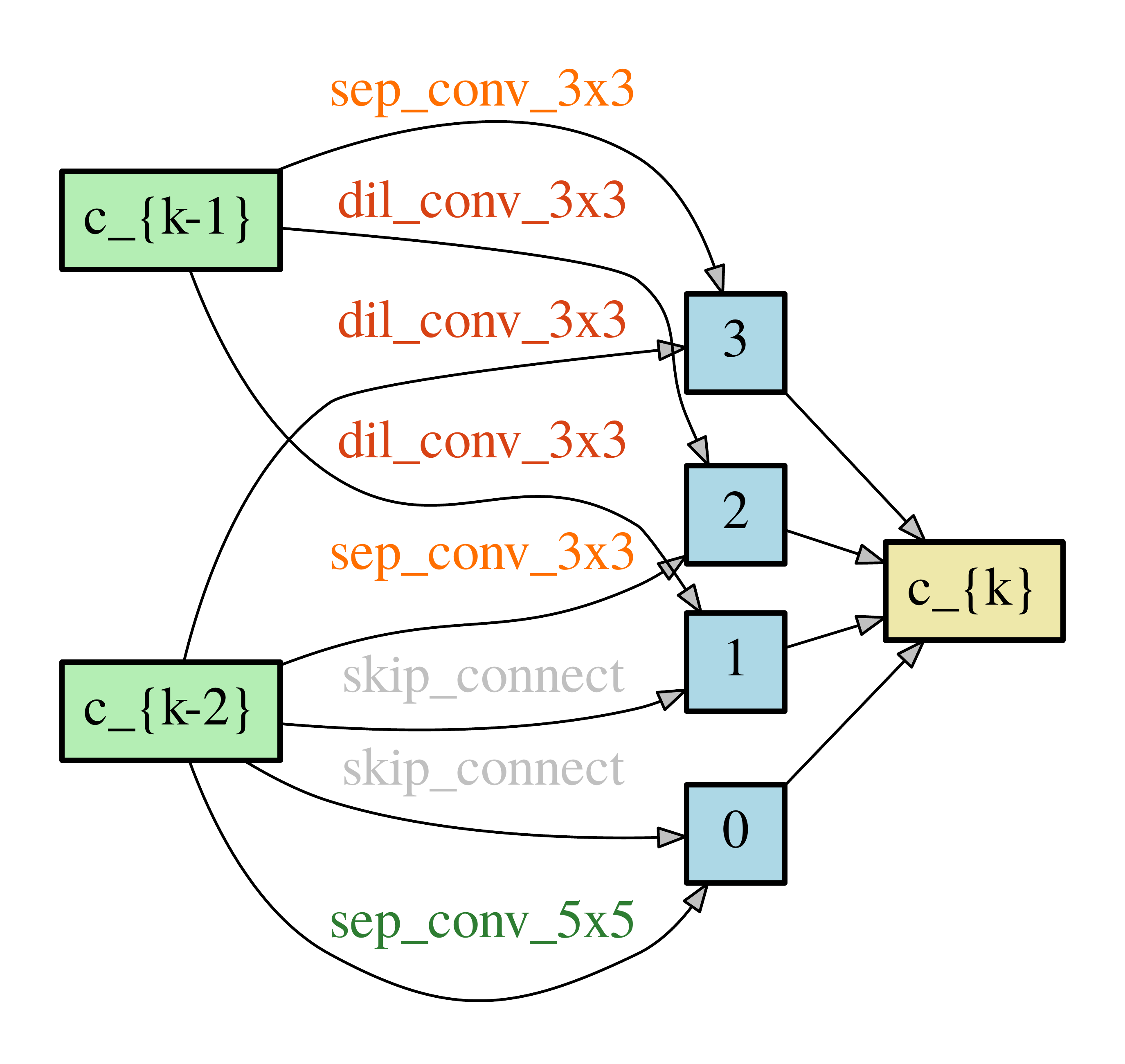}
    \end{subfigure}
    \begin{subfigure}{0.6\textwidth}
    \centering
        \caption{Reduce Cell}
        \label{fig:ap_cifar10_reduce}
        \includegraphics[scale=0.25]{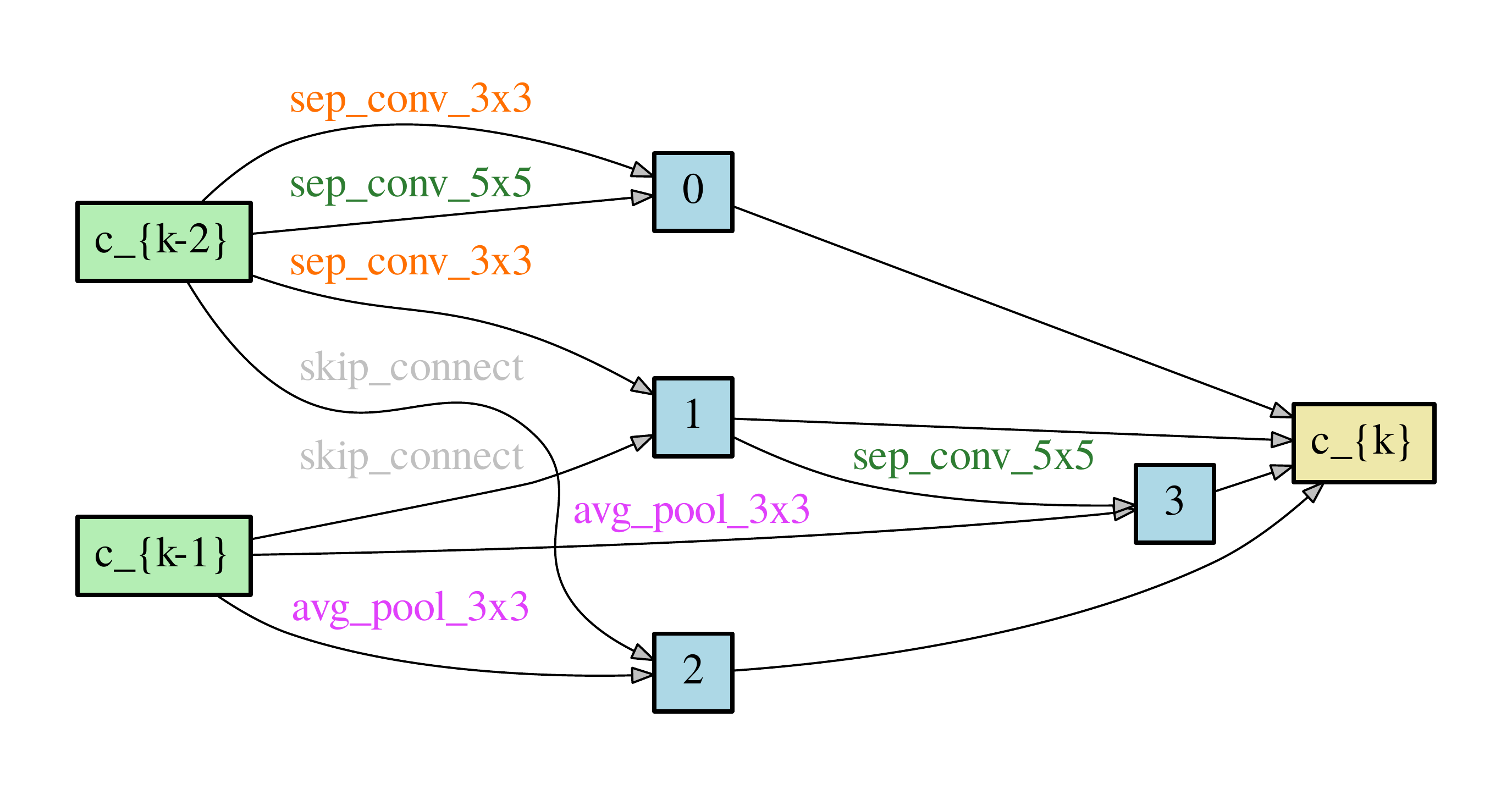}
    \end{subfigure}
\end{figure}

\begin{figure}[h]
    \vspace{-0.6cm}
    \setlength{\belowcaptionskip}{0pt}
    \centering
    \caption{The best performing cell found on CIFAR-100.}
    \label{fig:ap_cifar100_best_cell}
    \begin{subfigure}{.3\textwidth}
    \centering
        \caption{Normal Cell}
        \label{fig:ap_cifar100_normal}
        \includegraphics[scale=0.15]{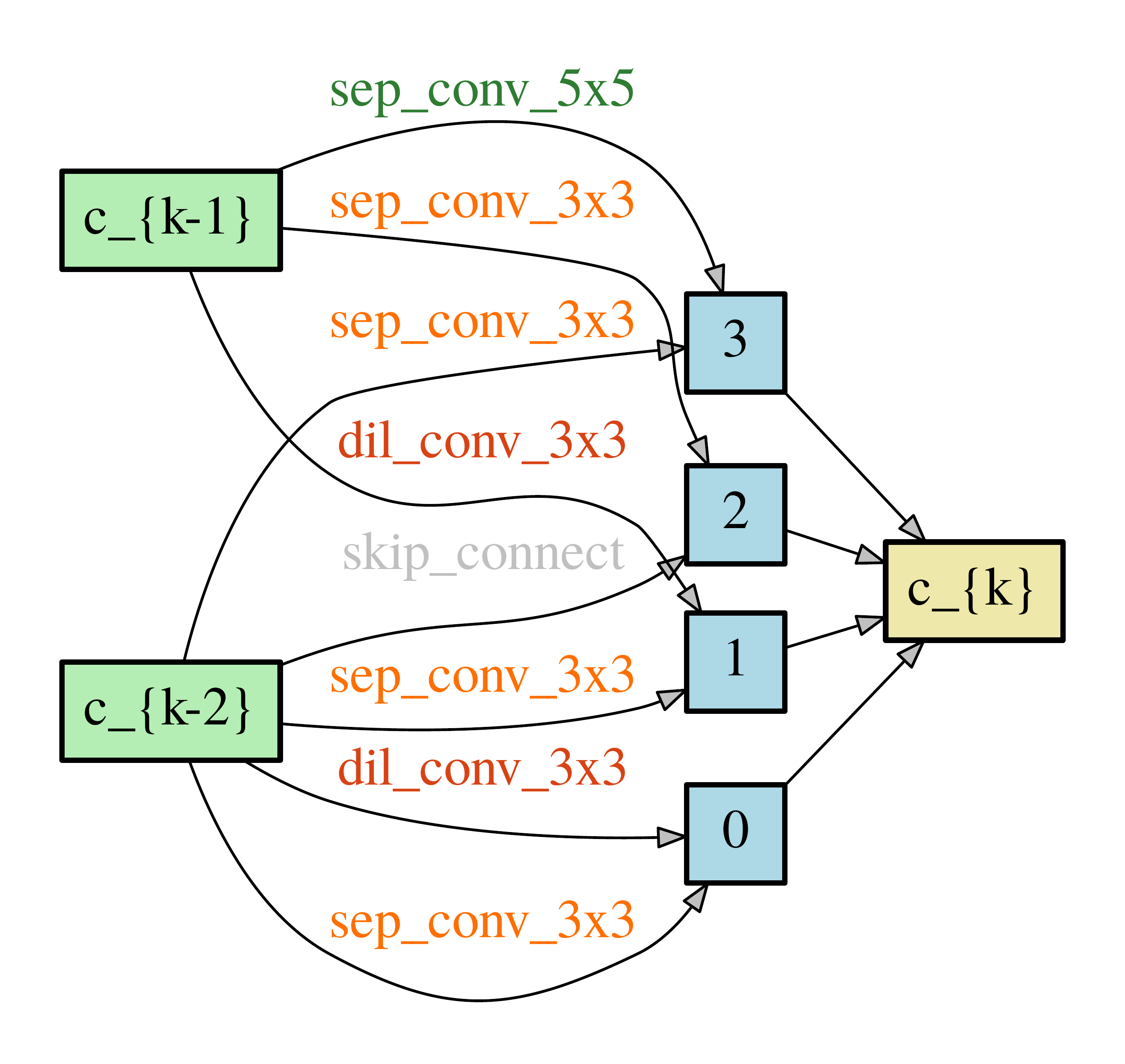}
    \end{subfigure}
    \begin{subfigure}{0.6\textwidth}
    \centering
        \caption{Reduce Cell}
        \label{fig:ap_cifar100_reduce}
        \includegraphics[scale=0.25]{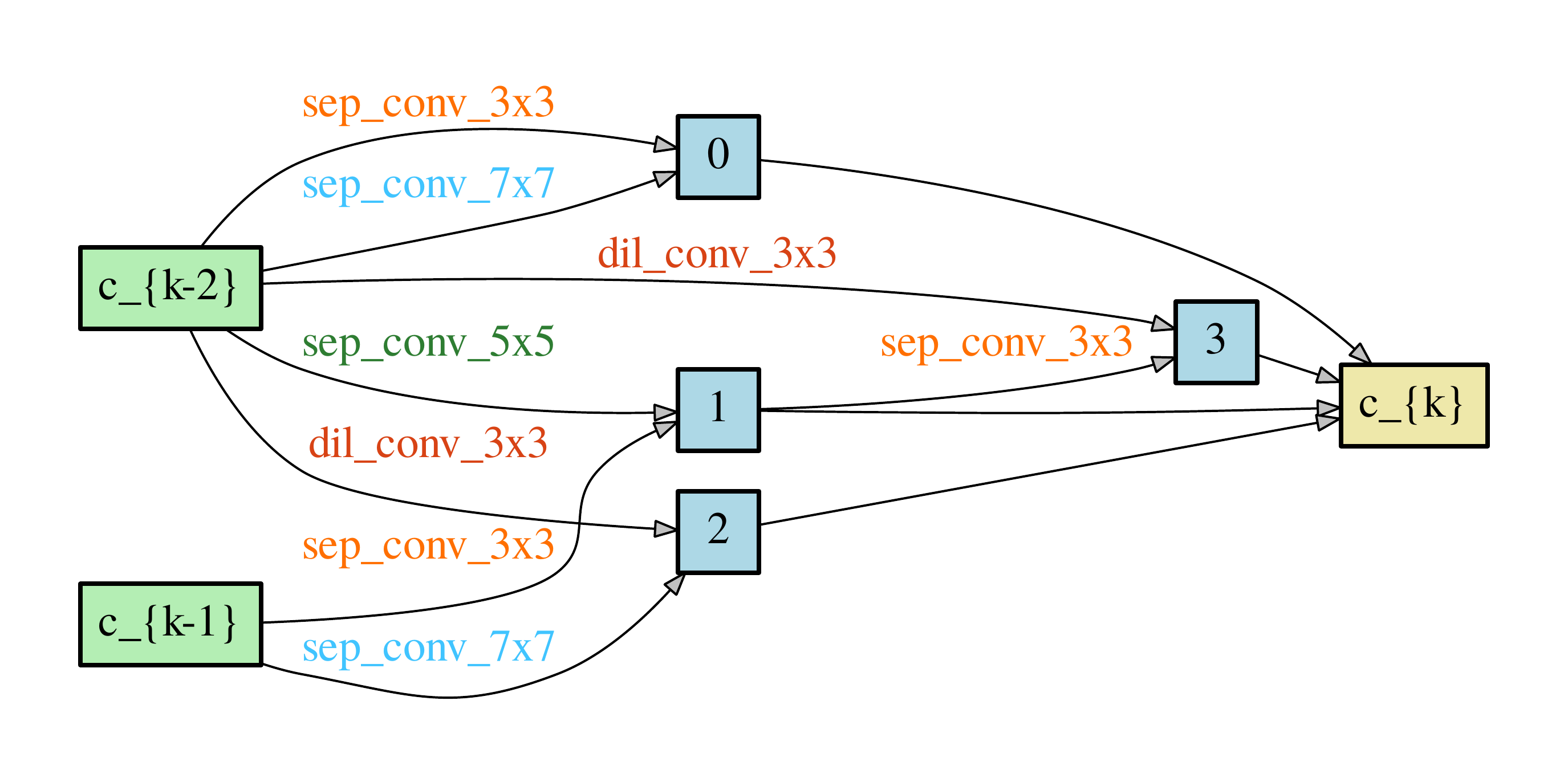}
    \end{subfigure}
\end{figure}

% \subsection{Best Random Cell}
\begin{figure}[t]
    \setlength{\belowcaptionskip}{0pt}
    \centering
    \caption{The best performing randomly proposed cell on ImageNet.}
    \label{fig:ap_random_best_cell}
    \begin{subfigure}{0.49\textwidth}
    \centering
        \caption{Normal Cell}
        \label{fig:ap_random_normal}
        \includegraphics[trim={0 0 0 1.5cm},clip,scale=0.25]{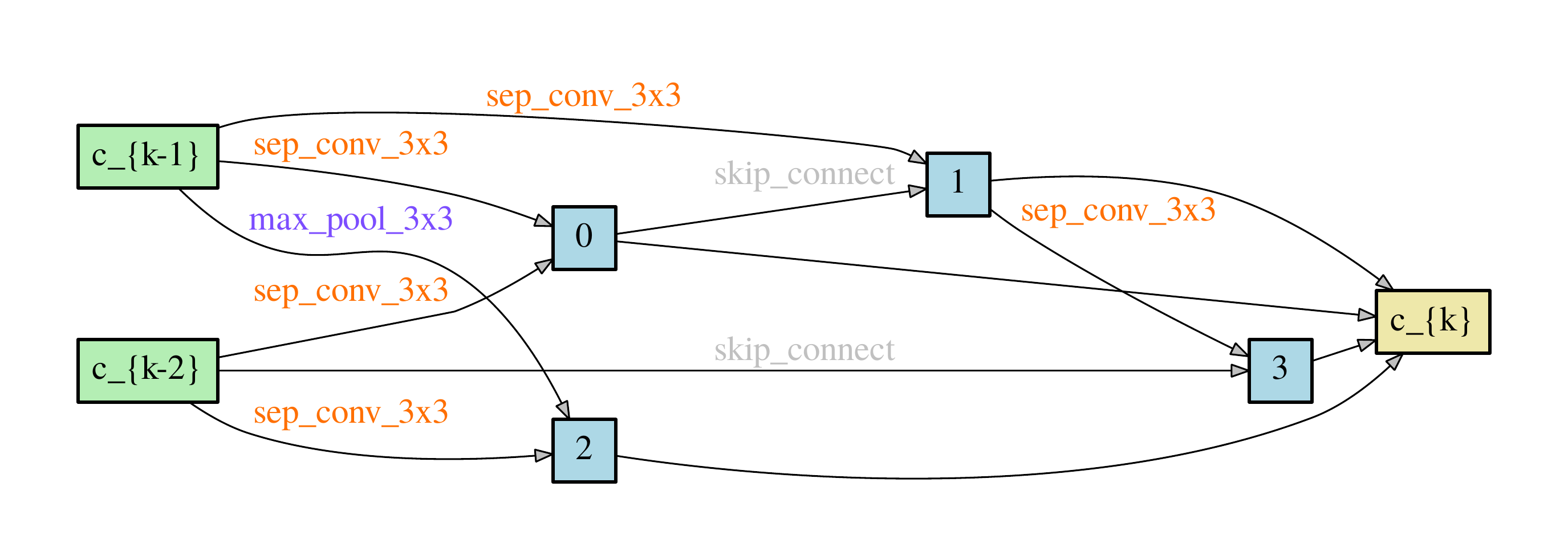}
    \end{subfigure}
    \begin{subfigure}{0.49\textwidth}
    \centering
        \caption{Reduce Cell}
        \label{fig:ap_random_reduce}
        \includegraphics[trim={0 0 0 1.5cm},clip,scale=0.25]{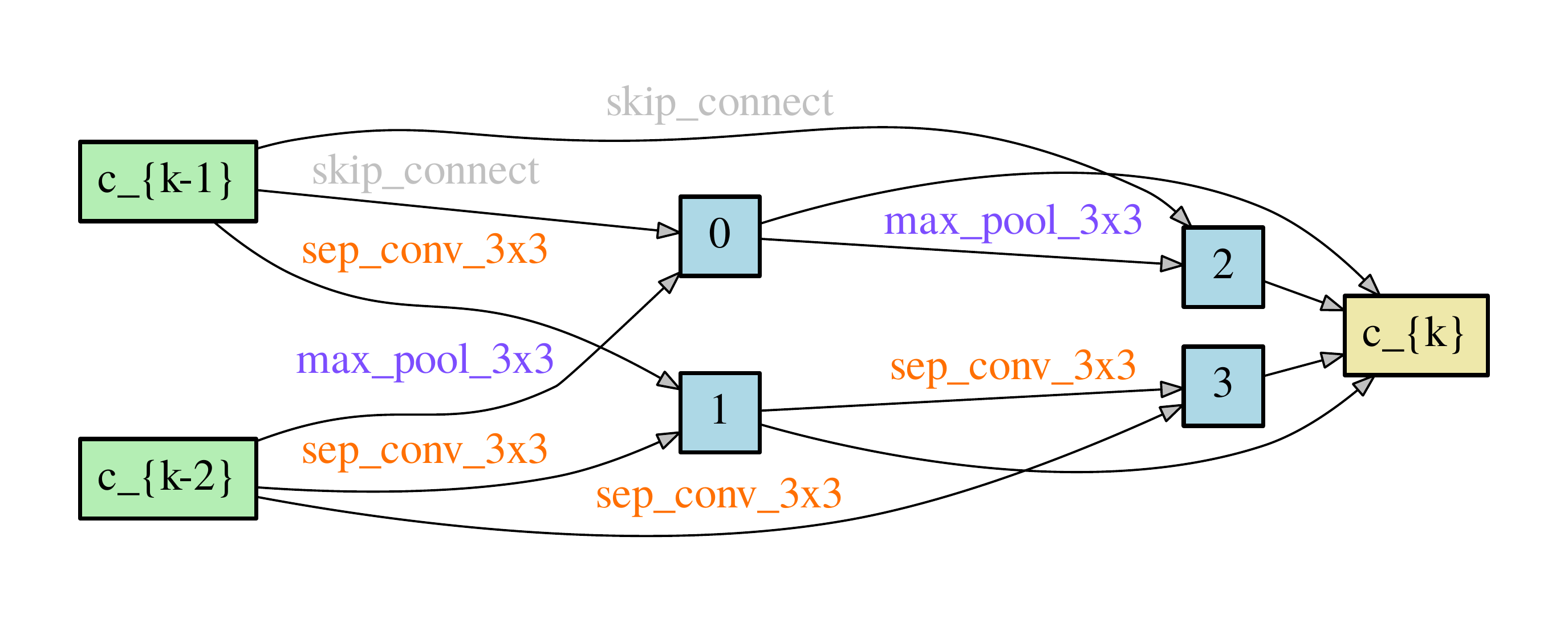}
    \end{subfigure}
\end{figure}

\clearpage

\section{Comparison with the Previous Work in DARTS Space}\label{app:previous}
In this section, we compare the best cells discovered by UNAS against previously published results on CIFAR-10, CIFAR-100 and ImageNet. 

\textbf{CIFAR-10:} In Table~\ref{table:results_cifar10}, the best cell discovered by UNAS is compared against the previous work that uses similar search space. For DARTS and P-DARTS, we list the original results reported by the authors, as well as, the best cell we discovered by running the original implementation four times. The best cell discovered by UNAS outperforms DARTS and SANS. In comparison to P-DARTS, UNAS obtains better than the best cell that we discovered by running the original P-DARTS code four times with different seeds. However, UNAS achieves a comparable result to P-DARTS' originally reported result on CIFAR-10. Nevertheless, as we show in Table~\ref{table:cifar10_ab}, UNAS outperforms DARTS, P-DARTS, and SNAS in terms of the average performance. As discussed by Li and Talwalkar~\cite{li2019random}, the average performance is a better representative metric to evaluate the performance of NAS methods, as it is more robust against rare architecture instances that perform well, but, are less likely to be discovered by the method. Such architectures require many search/evaluation runs, making NAS models expensive for practical applications, and more challenging for reproducing the results. 

When we ran the original P-DARTS source code with four different initialization seeds\footnote{We exactly followed the hyperparameters and commands using the search/eval code provided by the authors. We only set the initialization seed to a number in $\{0, 1, 2, 3\}$.}, we could not find an architecture with accuracy similar to the reported number. We believe this is because i) P-DARTS reports the lowest error observed during the evaluation phase while we report the error at the end of evaluation following DARTS. Taking the minimum of test error values, across small fluctuations towards the end of training, can reduce the error rate by 0.1\%, ii) P-DARTS does not report the number of searches performed to obtain the best result. We hypothesize that the reported result is the best architecture obtained from many searches. However, we do not intend to discount the contributions made by P-DARTS. When we evaluate the original discovered cell by P-DARTS on CIFAR-10, we can reproduce the same results in the evaluation phase. Nevertheless, the contributions of UNAS are orthogonal to P-DARTS thesis as discussed in Sec.~\ref{sec:related}. UNAS proposes new gradient estimators that work with differentiable and non-differentiable objective functions and it also introduces a new objective function based on the generalization gap.

\textbf{CIFAR-100:} In Table~\ref{table:results_cifar100}, our best cell discovered using UNAS is compared against previous work. We can see that UNAS outperforms DARTS, SANS, and P-DARTS on this dataset. Similar to CIFAR-10, when we ran P-DARTS code four times, we could not discover a cell as performant as the cell discovered originally on CIFAR-100.

\begin{table}[h]
    \setlength{\tabcolsep}{1pt}
    \setlength{\belowcaptionskip}{-5pt}
    \caption{Results on CIFAR-10.}
    \label{table:results_cifar10}
    \centering
    \resizebox{1.0\linewidth}{!}{%
        \begin{tabular}{lcccc}
            \toprule
            \multirow{2}{*}{\bf Architecture} & {\bf Test Error} & {\bf Params} & {\bf Search Cost} & {\bf Search} \\
            & (\%) & (M) & (GPU Days) & {\bf Method} \\
            \midrule
            % DenseNet-BC~\cite{huang2017densely}               & 3.46 & 25.6 & -- & -- & manual \\
            % \midrule
            NASNet-A ~\cite{zoph2018learning}        & 2.65 & 3.3 & 2000 & RL \\
            BlockQNN~\cite{zhong2018practical}       & 3.54 & 39.8 & 96 & RL \\
            AmoebaNet-A~\cite{real2018regularized}   & 3.12 & 3.1 & 3150 & evolution \\
            % \rowcolor{lightgray}
            AmoebaNet-B~\cite{real2018regularized}   & 2.55 & 2.8 & 3150 & evolution \\
            H. Evolution~\cite{liu2017hierarchical} & 3.75 & 15.7 & 300 & evolution \\
            PNAS~\cite{liu2018progressive}           & 3.41 & 3.2 & 225 & SMBO \\
            ENAS~\cite{pham2018efficient}            & 2.89 & 4.6 & 0.45 & RL \\
            Random~\cite{liu2018darts}               & 3.29 & 3.2 & 4 & random \\
            
            \midrule
            DARTS-1$^{st}$~\cite{liu2018darts}  & 3.00 & 3.3 & 1.5 & grad-based \\
            DARTS-2$^{nd}$~\cite{liu2018darts} & 2.76 & 3.3 & 4 & grad-based \\
            
            % SNAS (mild) + cutout~\cite{xie2018snas}           & 2.98 & 2.9 & 1.5 & 7 & grad-based \\
            SNAS~\cite{xie2018snas}       & 2.85 & 2.8 & 1.5 & grad-based \\
            P-DARTS~\cite{xie2018snas}       & 2.50 & 3.4 & 0.3 & grad-based \\
            %P-DARTS-Large~\cite{xie2018snas}       & 2.25 & 10.5 & 0.3 & 7 & grad-based \\
            % SNAS (aggressive) + cutout~\cite{xie2018snas}     & 3.10 $\pm$ 0.04 & 2.3 & 1.5 & 7 & grad-based \\
            % Proxyless-R + cutout~\cite{cai2018proxylessnas}   & 2.30 & 5.8 & -- & -- & grad-based \\
            % Proxyless-G + cutout~\cite{cai2018proxylessnas}   & 2.08 & 5.7 & -- & -- & grad-based \\
            \midrule
            \multicolumn{5}{c}{\textit{Best cell discovered after running the original code 4 times}} \\
            DARTS-2$^{nd}$~\cite{liu2018darts} & 2.80 & 3.6 & 4 & grad-based \\
            P-DARTS~\cite{xie2018snas}                 & 2.75 & 3.5 & 0.3 & grad-based \\
            %P-DARTS-Large~\cite{xie2018snas}           & ?? & ?? & ?? & grad-based \\
            \midrule
            UNAS                                &   2.53            &  3.3 &  4.3   & grad RL \\
            %\bf UNAS-Large                          &   \bf ?? $\pm$ ??         &  \bf 10.2 &  \bf 4.3  &   \bf 7  & \bf grad-based RL \\
            \bottomrule
            %\vspace{-0.8cm}
        \end{tabular} %
    }
% \vspace{-2cm}
\end{table}

\begin{table}[h]
    \setlength{\tabcolsep}{1pt}
    \setlength{\belowcaptionskip}{-5pt}
    \caption{Results on CIFAR-100.}
    \label{table:results_cifar100}
    \centering
    \resizebox{1.0\linewidth}{!}{%
        \begin{tabular}{lcccc}
            \toprule
            \multirow{2}{*}{\bf Architecture} & {\bf Test Error} & {\bf Params} & {\bf Search Cost} & {\bf Search} \\
            & (\%) & (M) & (GPU Days) & {\bf Method} \\
            \midrule
            % DenseNet-BC~\cite{huang2017densely}               & 3.46 & 25.6 & -- & -- & manual \\
            % \midrule
            BlockQNN~\cite{zhong2018practical}                & 18.06 & 39.8 & 96 & RL \\
            \midrule
            P-DARTS~\cite{xie2018snas}             & 15.92 & 3.6 & 0.3 & grad-based \\
            %P-DARTS-Large~\cite{xie2018snas}       & 14.64 & 11.0 & 0.3 & 7 & grad-based \\
            % SNAS (aggressive) + cutout~\cite{xie2018snas}     & 3.10 $\pm$ 0.04 & 2.3 & 1.5 & 7 & grad-based \\
            % Proxyless-R + cutout~\cite{cai2018proxylessnas}   & 2.30 & 5.8 & -- & -- & grad-based \\
            % Proxyless-G + cutout~\cite{cai2018proxylessnas}   & 2.08 & 5.7 & -- & -- & grad-based \\
            \midrule
            \multicolumn{5}{c}{\textit{Best cell discovered after running the original code 4 times}} \\
            DARTS-2$^{nd}$~\cite{liu2018darts} & 20.49 & 1.8 & 4 & grad-based \\
            P-DARTS~\cite{xie2018snas}                & 17.36 & 3.7 & 0.3 & grad-based \\
            %P-DARTS-Large~\cite{xie2018snas}           & ?? & ?? & ?? & 7 & grad-based \\
            \midrule
            UNAS                                &   15.79   &  4.1 & 4.0  & grad RL \\
            %\bf UNAS-Large                             &   \bf ?? $\pm$ ??         &  \bf 12.31 &  \bf 4.0  &   \bf 7  & \bf grad-based RL \\
            \bottomrule
            %\vspace{-0.8cm}
        \end{tabular} %
    }
% \vspace{-2cm}
\end{table}

\begin{table*}
    \setlength{\tabcolsep}{2pt}
    \caption{Best results on ImageNet in the mobile setting (\#Multi.-Adds$<$600M)~\cite{howard2017mobilenets}.}
    \label{table:results_imagenet}
    \centering
    \resizebox{0.7\linewidth}{!}{%
        \begin{tabular}{lcccccc}
            \toprule
            \multirow{2}{*}{\bf Architecture} & \multicolumn{2}{c}{\bf \underline{Val Error (\%)}} & {\bf Params} & {\bf $\boldsymbol{\times}\boldsymbol{+}$} & {\bf Search Cost} & \multirow{2}{*}{\bf Search Method}\\
            & top-1 & top-5 & (M) & (M) & (GPU Days) \\
            \midrule
            % MobileNet~\cite{howard2017mobilenets}               & 29.4 & 10.5 & 4.2 & 569 & -- & manual \\
            MobileNetV2~\cite{sandler2018mobilenetv2}          & 25.3 & --   & 6.9 & 585 & -- & manual \\
            % ShuffleNet 2$\times$(v1)~\cite{zhang2018shufflenet} & 26.3 & 8.5 & 5.4 & 524 & -- & manual \\
            % \rowcolor{lightgray}
            ShuffleNetV2 2$\times$~\cite{ma2018shufflenet}    & 25.1 & 7.8 & 7.4 & 591 & -- & manual \\
            \midrule
            NASNet-A~\cite{zoph2018learning}       & 26.0 & 8.4 & 5.3 & 564 & 2000 & RL \\
            % NASNet-B~\cite{zoph2018learning}       & 27.2 & 8.7 & 5.3 & 488 & 2000 & RL \\
            % NASNet-C~\cite{zoph2018learning}       & 27.5 & 9.0 & 4.9 & 558 & 2000 & RL \\
            % AmoebaNet-A~\cite{real2018regularized} & 25.5 & 8.0 & 5.1 & 555 & 3150 & evolution \\
            AmoebaNet-B~\cite{real2018regularized} & 26.0 & 8.5 & 5.3 & 555 & 3150 & evolution \\
            % \rowcolor{lightgray}
             AmoebaNet-C~\cite{real2018regularized} & 24.3 & 7.6 & 6.4 &  570& 3150 & evolution \\
            PNAS~\cite{liu2018progressive}         & 25.8 & 8.1 & 5.1 & 588 & $\sim$255 & SMBO \\
            % \midrule
            DARTS~\cite{liu2018darts}              & 26.7 & 8.7 & 4.7 & 574 & 4   & grad-based \\
            SNAS~\cite{xie2018snas}                & 27.3 & 9.2 & 4.3 & 522 & 1.5 & grad-based \\
            P-DARTS~\cite{chen2019pdarts}          & 24.4 & 7.4 & 4.9 & 557 & 0.3 & grad-based \\
            \midrule
            \multicolumn{7}{c}{\textit{Best cell discovered after running the original code 4 times}} \\
            DARTS~\cite{liu2018darts}              & 25.2 & 7.7 & 5.12 & 595 & 4   & grad-based \\
            P-DARTS~\cite{chen2019pdarts}          & 24.5 & 7.3 & 5.2 & 599 & 0.3 & grad-based \\
            \midrule
            Random Cell                      & 25.55 & 8.06 & 5.37 & 598 & $\sim\!250$ & random \\
            %Ours - CIFAR-10                        & 25.14 & 7.82 & 5.13 & 574.61 & 4.5 & grad-based RL \\
            % \rowstyle{\bfseries}
            UNAS                  &  24.46 &  7.44 & 5.07 & 563 & 16 & grad-based RL \\
            \bottomrule
            % \vspace{-34pt}
        \end{tabular}
    }
% \vspace{-25pt}
\end{table*}

% \subsection{Architecture Search on ImageNet}
\textbf{ImageNet:}
Here, we compare UNAS on the ImageNet dataset against previous works. We also provide a surprisingly strong baseline using randomly generated architectures. Table~\ref{table:results_imagenet} summarizes the results.

{\bf Random Baseline:} We provide a strong random baseline, indicated by ``Random Cell'' in Table~\ref{table:results_imagenet}, that outperforms most prior NAS methods.
Random cells are generated by drawing uniform random samples from factorized cell structure.
%Each node randomly chooses two nodes as input (without replacement) from the available inputs (\ie two previous layers and the previous nodes in the cell), and also sample operations on those edges randomly (with replacement). 
We train a total of 10 networks constructed by randomly generated Normal and Reduce cells. The best network yields top-1 and top-5 errors of 25.55\% and 8.06\% respectively (see Fig~\ref{fig:ap_random_best_cell} for the cell structure).
To the best of our knowledge, we are the first to report performance of a randomly discovered cell on ImageNet that outperforms most previous NAS methods, although not UNAS and P-DARTS.
% The high performance of this baseline indicates that the search space is of paramount importance.

{\bf Direct Search on ImageNet:}
Searching on ImageNet gives us the cell in Fig.~\ref{fig:ap_imagenet_best_cell}.
Our cell searched on ImageNet obtains a performance, comparable to P-DARTS and AmoebaNet-C~\cite{real2018regularized}, giving a top-1 and top-5 error of 24.46\% and 7.44\% resp.\ at a fraction of the cost (0.5\%) required by the best AmoebaNet-C~\cite{real2018regularized}.

%%%%%%%%%%%%%%%%%%%%%%%%%%%%%%%%

\section{UNAS with ProxylessNAS Search Space}\label{app:latency_settings}
In this section, we list the implementation details used for the latency based experiments presented in Sec.~\ref{subsec:latency}.

\subsection{Search Space}
We follow ProxylessNAS~\cite{cai2018proxylessnas} to construct the search space which is based on MobileNetV2~\cite{sandler2018mobilenetv2}. During search we seek operations assigned to each layer of a 21-layer network. The operations in each layer are constructed using mobile inverted residual blocks~\cite{sandler2018mobilenetv2} by varying the kernel size in $\{3, 5, 7\}$ and the expansion ratio in $\{3, 6\}$ yielding 6 choices with the addition of a skip connection (i.e., an identity operation) which enables removing layers. For the channel sizes, we followed the ProxylessNAS-GPU architecture. 
For the first 20 epochs, only the network parameters ($\w$) are trained, while the architecture parameters ($\bphi$) are frozen. The architecture parameters are trained in 15 epochs using the Adam optimizer with cosine learning rate schedule starting from 1$\times10^{-3}$ annealed down to 3$\times10^{-4}$. The network parameters are also trained using Adam with cosine learning rate schedule starting from 3$\times10^{-4}$ annealed down to 1$\times10^{-4}$. Batch of 192 images on 8 V-100 GPUs are used for training. For the latency-based search, we use the following objective function:
\begin{equation}
    \E_{p_{\bphi}(\z)}[\L_{\gen}(\z, \w)] + \lambda_{lat} \E_{p_{\bphi}(\z)}[f(\L_{lat}(\z) - t_\text{target})] \nonumber
\end{equation}
where $t_\text{target}$ represents the target latency, $f(u)=max(0, u)$ penalizes the architectures that has latency higher than the target latency.

We linearly anneal $\lambda_{lat}$ from zero to 0.1 to focus the architecture search on the classification loss initially. However, we empirically observed that the latency loss has a low gradient variance that provides a very strong training signal for selecting low-latency operations such as skip connection. To avoid this, Inspired by P-DARTS~\cite{chen2019pdarts}, we apply dropout to the skip connection during search. We observe that a small amount of dropout with probability 0.1 prevents the search from over-selecting the skip operation. 

\subsection{Evaluation}
After search, the operations in each layer with the highest probability values are chosen for the final network. The training in the evaluation phase is based on the ProxylessNAS evaluation. Batches of 512 images on 8 V-100 GPUs are used for training in 300 epochs. We train our networks using SGD with momentum of 0.9, base learning rate of 0.2, linear learning-rate warmup in 5 epoch, and weight decay of 5$\times10^{-5}$. The learning rate is annealed to 0 throughout the training using a cosine learning rate decay.

\end{document}